\documentclass[10pt,twocolumn,letterpaper]{article}

\usepackage{iccv}
\usepackage{times}
\usepackage{epsfig}
\usepackage{graphicx}
\usepackage{amsmath}
\usepackage{amssymb}

\usepackage[pagebackref=true,breaklinks=true,colorlinks,bookmarks=false]{hyperref}

\iccvfinalcopy 

\def\iccvPaperID{5429} 

\ificcvfinal\fi

\usepackage{graphicx}
\usepackage{amsmath}
\usepackage{amssymb}
\usepackage{booktabs}

\usepackage{color}
\usepackage{makecell}
\newcolumntype{Y}{>{\centering\arraybackslash}X}

\usepackage{tikz}
\usepackage{pgfplots}
\usepackage{graphicx}
\usepackage{amsmath}
\usepackage{amssymb}
\usepackage{booktabs}
\usepackage{url}
\usepackage{epsfig}
\usepackage{lipsum}
\usepackage{amsthm}
\usepackage{comment}
\usepackage{multirow}
\usepackage{subcaption}
\usepackage{microtype}
\usepackage{xspace}
\usepackage{enumitem}
\usepackage{graphbox}
\usepackage{epigraph}
\usepackage{wrapfig}
\usepackage{textcomp}  
\usepackage{scalerel}  
\usepackage{animate}

\pgfplotsset{xtick style={draw=none}}
\pgfplotsset{ytick style={draw=none}}
\pgfplotsset{major grid style={gray!40}}
\pgfplotsset{every axis plot/.style={thick, mark size=1.0pt}}
\pgfplotsset{legend image code/.code={\draw[mark repeat=2, mark phase=2] plot coordinates {(0cm, 0cm) (0.2cm, 0cm) (0.4cm, 0cm)};}} 
\definecolor{C0}{rgb}{0.121569, 0.466667, 0.705882}
\definecolor{C1}{rgb}{1.000000, 0.498039, 0.054902}
\definecolor{C2}{rgb}{0.172549, 0.627451, 0.172549}
\definecolor{C3}{rgb}{0.839216, 0.152941, 0.156863}
\definecolor{C4}{rgb}{0.580392, 0.403922, 0.741176}
\newcommand{\plothh}{}\newcommand{\plothc}{}\newcommand{\plotvv}{}
\newcommand{\tickYtopD}[1]{\raisebox{-1.5ex}[0ex][0ex]{#1}}
\newcommand{\tickFVD}{\tickYtopD{FVD}}

\newcommand{\topic}[1]
{
\vspace{2mm}\noindent\textbf{#1}
}

\usepackage{xspace}
\usepackage{animate}
\usepackage{colortbl}
\usepackage{tabularx,booktabs}

\usepackage[section]{placeins}
\usepackage{float}

\usepackage[capitalize]{cleveref}
\crefname{section}{Sec.}{Secs.}
\Crefname{section}{Section}{Sections}
\Crefname{table}{Table}{Tables}
\crefname{table}{Tab.}{Tabs.}

\newcommand{\xx}{\mathbf{x}}
\newcommand{\ee}{\mathbf{e}}
\newcommand{\bzero}{\mathbf{0}}
\newcommand{\bI}{\mathbf{I}}
\newcommand{\cN}{\mathcal{N}}
\newcommand{\beps}{\mathbf{\epsilon}}
\newcommand{\ediffi}{eDiff-I\xspace}
\newcommand{\ediffivideo}{PYoCo\xspace}

\newcommand{\new}[1]{{#1}}

\newcommand{\tabcspace}{\vspace{-2mm}}
\newcommand{\tabspace}{\vspace{-4mm}}

\begin{document}

\title{Preserve Your Own Correlation: A Noise Prior for Video Diffusion Models}


\author{
Songwei Ge\thanks{Work done during the internship at NVIDIA.} \\
\small University of Maryland \\
\and
Seungjun Nah \\
\small NVIDIA \\
\and
Guilin Liu \\
\small NVIDIA \\
\and
Tyler Poon \\
\small University of Chicago \\
\and
Andrew Tao \\
\small NVIDIA \\
\and
Bryan Catanzaro \\
\small NVIDIA \\
\and
David Jacobs \\
\small University of Maryland \\
\and
Jia-Bin Huang \\
\small University of Maryland \\
\and
Ming-Yu Liu \\
\small NVIDIA \\
\and
Yogesh Balaji \\
\small NVIDIA \\
}

\twocolumn[{%
\maketitle
\renewcommand\twocolumn[1][]{#1}%

\vspace{-7mm}
\setlength{\tabcolsep}{0.5pt}
\renewcommand{\arraystretch}{0.5}
\begin{tabular}{c c c}
    \animategraphics[width=0.33\textwidth]{16}{videos/teaser_512/panda/}{0001}{0076} & 
    \animategraphics[width=0.33\textwidth]{16}{videos/teaser_512/supernova/}{0001}{0076} &
    \animategraphics[width=0.33\textwidth]{16}{videos/teaser_512/bear/}{0001}{0076} 
\end{tabular}
\begin{tabularx}{\textwidth}{m{0.318\textwidth} c m{0.318\textwidth} c m{0.318\textwidth}} 
    \emph{\small A very happy fuzzy panda dressed as a chef eating pizza in the New York street food truck.} & \hspace{0.6em} & 
    \emph{\small The supernova explosion of a white dwarf in the universe, photo realistic.} & \hspace{0.6em} &
    \emph{\small  A high-quality 3D render of hyperrealist, super strong, multicolor stripped, and fluffy bear with wings, highly detailed.} 
\end{tabularx}
\captionof{figure}{Given a text description, our approach can faithfully generate videos that are consistent with the input text while being photorealistic and temporally consistent. \emph{Best viewed with Acrobat Reader. Click the images to play the video clips.}}
\vspace{5mm}
\label{fig:teaser_512}

}]

\renewcommand{\thefootnote}{\fnsymbol{footnote}}
\footnotetext[1]{Work done during an internship at NVIDIA.}


\begin{abstract}
Despite tremendous progress in generating high-quality images using diffusion models, synthesizing a sequence of animated frames that are both photorealistic and temporally coherent is still in its infancy. 
While off-the-shelf billion-scale datasets for image generation are available, collecting similar video data of the same scale is still challenging. 
Also, training a video diffusion model is computationally much more expensive than its image counterpart. 
In this work, we explore finetuning a pretrained image diffusion model with video data as a practical solution for the video synthesis task.
We find that naively extending the image noise prior to video noise prior in video diffusion leads to sub-optimal performance. Our carefully designed video noise prior leads to substantially better performance.
Extensive experimental validation shows that our model, Preserve Your Own COrrelation (\ediffivideo), attains SOTA zero-shot text-to-video results on the UCF-101 and MSR-VTT benchmarks. It also achieves SOTA video generation quality on the small-scale UCF-101 benchmark with a $10\times$ smaller model using significantly less computation than the prior art. 
The project page is available at \url{https://research.nvidia.com/labs/dir/pyoco/}.
\end{abstract}

\section{Introduction}\label{sec:intro}

\setlength{\tabcolsep}{0.5pt}
\renewcommand{\arraystretch}{0.5}
\begin{figure*}[t]
    \centering
    \subfloat[\label{fig:tsne_a}]{\includegraphics[height=5.1cm]{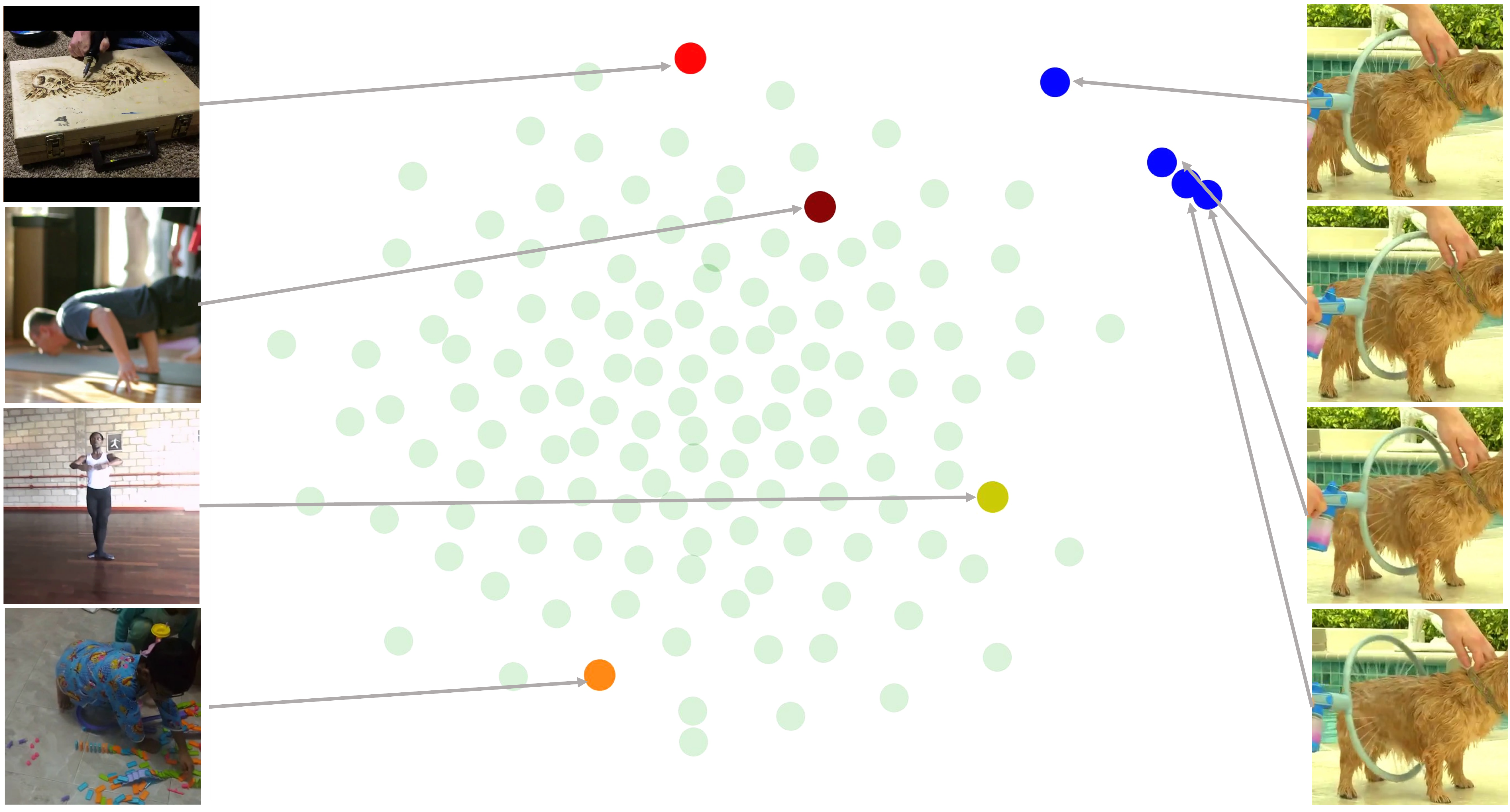}}
    \hfill
    \vline
    \hfill
    \subfloat[\label{fig:tsne_b}]{\includegraphics[height=5.1cm]{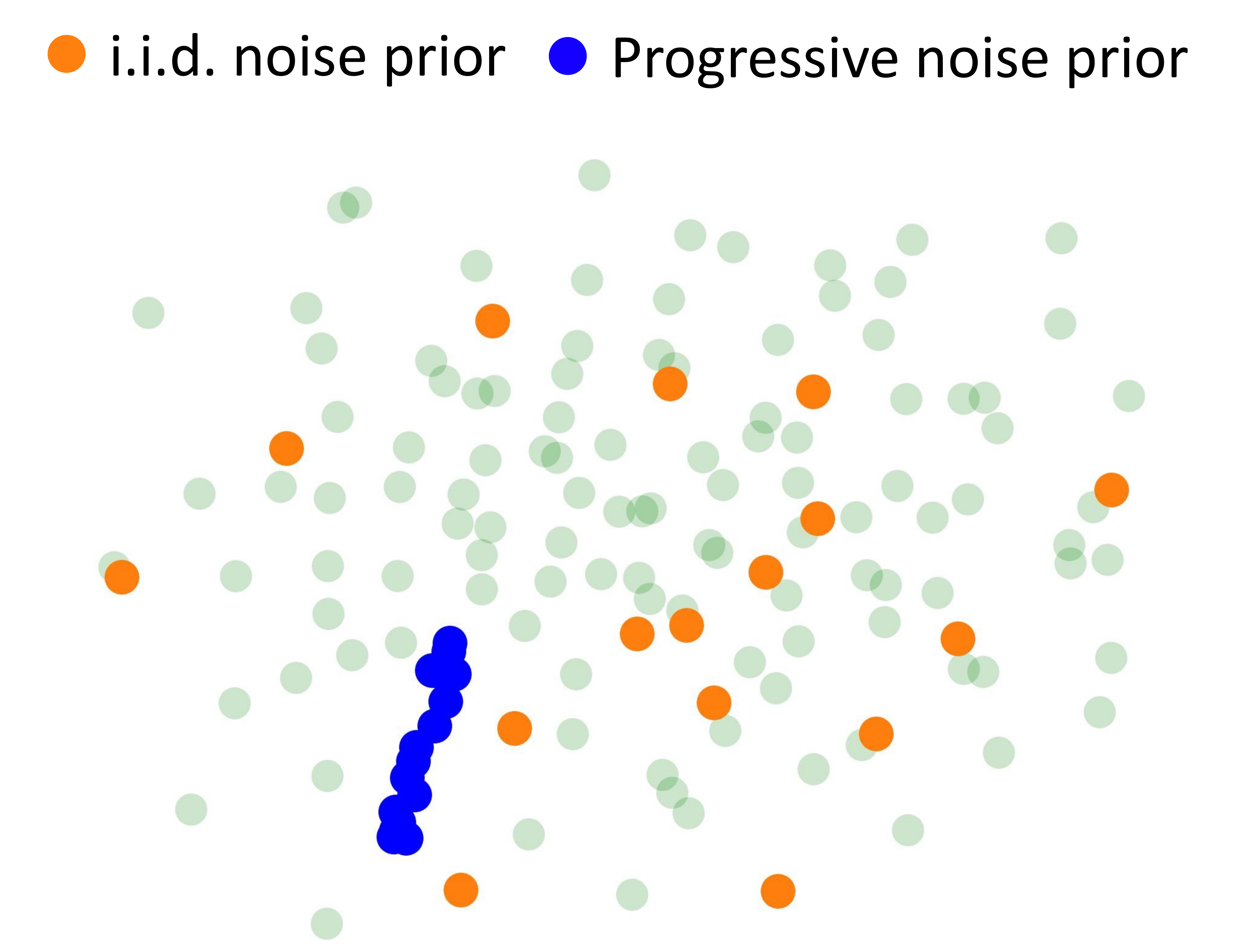}}
    \caption{\textbf{Visualizing the noise map correlations}. (a) visualizes the t-SNE plot of the noise maps corresponding to input frames randomly sampled from videos. These noise maps are obtained by running a diffusion ODE~\cite{song2021scorebased, song2021denoising} on the input frames using a trained text-to-image model, but in the opposite direction of image synthesis ($\sigma: 0 \rightarrow \sigma_{\text{max}}$). The green dots in the background  denote the reference noise maps sampled from an i.i.d. Gaussian distribution. The red dots and yellow dots are noise maps corresponding to input frames coming from different videos. We found they are spread out and share no correlation. On the other hand, the noise maps corresponding to the frames coming from the same video (shown in blue dots) are clustered together. (b) Using an i.i.d. noise model (orange dots) for finetuning text-to-image models for video synthesis is not ideal since temporal correlations between frames are not modeled. To remedy this, we propose a progressive noise model in which the correlation between different noise maps is injected along the temporal axis. Our progressive noise model (blue dots) aptly models the correlations present in the video noise maps.}
    \label{fig:tsne}
\end{figure*}

Large-scale diffusion-based text-to-image models~\cite{ramesh2022hierarchical,saharia2022photorealistic,balaji2022ediffi} have demonstrated impressive capabilities in turning complex text descriptions into photorealistic images. They can generate images with novel concepts unseen during training. Sophisticated image editing and processing tasks can easily be accomplished through guidance control and embedding techniques. Due to the immense success in several applications~\cite{meng2022sdedit,zhang2023adding,brooks2022instructpix2pix}, these models are established as powerful image synthesis tools for content generation. As image synthesis is largely democratized with the success of these text-to-image models, it is natural to ask whether we can repeat the same success in video synthesis with large-scale diffusion-based text-to-video models.

Multiple attempts have been made to build large-scale video diffusion models. Ho~\etal~\cite{ho2022video} proposed a UNet-based architecture for the video synthesis task that is trained using joint image-video denoising losses. Imagen video~\cite{ho2022imagen} extends the cascaded text-to-image generation architecture of Imagen~\cite{saharia2022photorealistic} for video generation. In both works, the authors directly train a video generation model from scratch. While these approaches achieve great success and produce high-quality videos, they are inherently expensive to train, requiring hundreds of high-end GPUs or TPUs and several weeks of training. After all, video generators not only need to learn to form individual images but should also learn to synthesize coherent temporal dynamics, which makes the video generation task much more challenging. While the formation of individual frames is a shared component in an image and video synthesis, these works disregard the existence of powerful pretrained text-to-image diffusion models and train their video generators from scratch.

We explore a different avenue for building large-scale text-to-video diffusion models by starting with a pretrained text-to-image diffusion model. Our motivation is that most of the components learned for the image synthesis task can effectively be reused for video generation, leading to knowledge transfer and efficient training. A similar idea is adopted by several recent works~\cite{singer2022make,zhou2022magicvideo,blattmann2023videoldm}.
Without exception, when finetuning, they naively extend the image diffusion noise prior (i.i.d. noise) used in the text-to-image model to a video diffusion noise prior by adding an extra dimension to the 2D noise map. We argue that this approach is not ideal as it does not utilize the natural correlations in videos that are already learned by the image models. This is illustrated in Figure~\ref{fig:tsne}, where we visualize the t-SNE plot of noise maps corresponding to different input frames as obtained from a pretrained text-to-image diffusion model. The noise maps corresponding to different frames coming from the same video (blue dots in Figure~\ref{fig:tsne_a} are clustered together, exhibiting a high degree of correlation. The use of i.i.d. noise prior does not model this correlation, which would impede the finetuning process. Our careful analysis of the video diffusion noise prior leads us to a noise prior that is better tailored for finetuning an image synthesis model to the video generation task. As illustrated in Figure~\ref{fig:tsne_b}, our proposed noise prior (shown in blue dots) aptly captures the correlations in noise maps corresponding to video frames.

We then proceed to build a large-scale diffusion-based text-to-video model. We leverage several design choices from the prior works, including the use of temporal attention~\cite{ho2022video}, joint image-video finetuning~\cite{ho2022video}, a cascaded generation architecture~\cite{ho2022imagen}, and an ensemble of
expert denoisers~\cite{balaji2022ediffi}. Together with these techniques and the proposed video noise prior, our model establishes a new state-of-the-art for video generation outperforming competing methods on several benchmark datasets. Figure~\ref{fig:teaser_512} shows our model can achieve high-quality zero-shot video synthesis capability with SOTA photorealism and temporal consistency.

In short, our work makes the following key contributions.

\begin{enumerate}[leftmargin=8pt]
    \setlength\itemsep{0em}
    \item We propose a video diffusion noise tailored for finetuning text-to-image diffusion models for text-to-video.
    \item We conduct extensive experimental validation and verify the effectiveness of the proposed noise prior.
    \item We build a large-scale text-to-video diffusion model by finetuning a pretrained \ediffi~model with our noise prior and achieve state-of-the-art results on several benchmarks.
\end{enumerate}

\section{Related Work}\label{sec:related_work}

\textbf{Diffusion-based text-to-image models: } Diffusion models have significantly advanced the progress of text-based photorealistic, compositional image generation~\cite{ramesh2022hierarchical,saharia2022photorealistic}. Given the nature of the iterative denoising process that requires massive numbers of score function evaluations, earlier diffusion models focused on generating low-resolution images, e.g., $64\times64$~\cite{ho2020denoising,song2021denoising}. To generate high-resolution images, two common approaches have been used. The first approach applies cascaded super-resolution models in the RGB space~\cite{nichol2021glide,ho2022cascaded,saharia2022photorealistic,ramesh2022hierarchical}, while the second approach leverages a decoder to exploit latent space~\cite{rombach2022high,gu2022vector}. Based on these models, advanced image and video editing have been achieved through finetuning the model~\cite{ruiz2022dreambooth,zhang2023adding,brooks2022instructpix2pix,kumari2022multi,wu2022tune,ma2023follow} or controlling the inference process~\cite{meng2022sdedit,hertz2022prompt,parmar2023zero,ge2023expressive,qi2023fatezero,ceylan2023pix2video,molad2023dreamix,bar2022text2live}. Here, we study the problem of using large-scale diffusion models for text-to-video generation. 

\textbf{Video generation models: }
Generating realistic and novel videos have long been an attractive and essential research direction~\cite{vondrick2016generating,ranzato2014video,yu2022magvit}. Previously studies have resorted to different types of generative models such as GANs~\cite{vondrick2016generating,saito2017temporal,Tulyakov_2018_CVPR,tian2021a,Shen_2023_CVPR}, Autoregressive models~\cite{srivastava2015unsupervised,yan2021videogpt,le2021ccvs,ge2022long,hong2022cogvideo}, and implicit neural representations~\cite{skorokhodov2021stylegan,yu2021generating}. Recently, driven by the tremendous success of applying the diffusion model to image synthesis, multiple works have proposed to explore diffusion models for conditional and unconditional video synthesis~\cite{voleti2022masked,harvey2022flexible,zhou2022magicvideo,wu2022tune,blattmann2023videoldm,khachatryan2023text2video,hoppe2022diffusion,voleti2022masked,yang2022diffusion,nikankin2022sinfusion,luo2023videofusion,an2023latent,wang2023videofactory}. For example, Singer~\etal extend the unCLIP framework~\cite{ramesh2022hierarchical} to text-to-video generation, which allows training without video captions~\cite{singer2022make}. Ho~\etal~\cite{ho2022video} extend the Imagen framework~\cite{saharia2022photorealistic} by repeatedly up-scaling low-resolution small-fps videos in both spatial and temporal directions with multiple models~\cite{ho2022imagen}. Our work also falls into this line of work which uses a diffusion model. We focus on augmenting an image diffusion model for video and study the design choice of the diffusion noise priors for such an image-to-video finetuning task.

\textbf{Leverage knowledge from images for text-to-video generation: } 
Like text-to-image models, text-to-video models require massive amounts of data to learn caption-relatedness, frame photorealism, and temporal dynamics. But in contrast to the abundant image data resource, video data are more limited in style, volume, and quality. To resolve such scarcity issue of text-video data, previous works have resorted to different strategies to leverage knowledge from image data for text-to-video generation, including joint training on the text-image data from scratch~\cite{ho2022video,ho2022imagen,villegas2022phenaki,wu2022nuwa}, first training a text-to-image model and then finetuning partially~\cite{hong2022cogvideo,blattmann2023videoldm,wu2022tune,ma2023follow} or entirely~\cite{singer2022make,esser2023structure} on the video dataset, and using CLIP image features as the conditional information~\cite{singer2022make,zhou2022magicvideo}. In this paper, we propose a new video diffusion noise prior that is tailored for finetuning a pretrained diffusion-based image generation model for the video generation task. We reuse several design choices in the prior work by finetuning jointly on text-image and text-video datasets. As a result, we can build a text-to-video generation system that achieves state-of-the-art zero-shot performances.

\section{Preliminaries}

Diffusion models generate data by iteratively denoising samples drawn from a noise distribution. In the case of text-to-video models, text embeddings obtained from a pre-trained text encoder are used as additional inputs in the denoising process. Formally, let $D(\xx, \ee, \sigma)$ denote a denoising network that operates on the noisy input video $\xx \in \mathbb{R}^{b\times n_{s} \times 3 \times h \times w}$ where $\ee$ is the text embedding, and $\sigma$ is the noise level. Here $n_{s}$ is the sequence length of the input video, $b$ is the batch size, and $h \times w$ is the spatial resolution. The model $D$ is trained to denoise the input $\xx$. 

\paragraph{Training}
We follow the EDM formulation of Karras~\etal\cite{karras2022elucidating} to optimize the denoiser $D$ using the following objective
\begin{align}\label{eq:denoising_objective}
    \mathbb{E}_{p_{\text{data}}(\xx_{\text{clean}}, \ee), p(\beps), p(\sigma)} & \left[\lambda(\sigma)\Vert D(\xx_{\text{noise}};\ee, \sigma) - \xx_{\text{clean}} \Vert^2_2 \right] \\
    & \text{where }~~ \xx_{\text{noise}} = \xx_{\text{clean}} + \sigma \beps \nonumber
\end{align}
Here, $\xx_{\text{noise}}$ is the noisy sample obtained by corrupting the clean video $\xx$ with noise $\sigma \beps$, where $p(\beps) = \cN(\bzero, \bI)$ and $\sigma$ is a scalar for the noise level drawn from $p(\sigma)$. The loss weight, $\lambda(\sigma)$, is a function of $\sigma$ given by $\lambda(\sigma) = (\sigma^2 + \sigma_{\text{data}}^2) / (\sigma \cdot \sigma_{\text{data}})^2$.
Eq.~\eqref{eq:denoising_objective} is a simple denoising objective in which the denoiser $D$ is trained to estimate the clean video $\xx_{\text{clean}}$ from the noisy input $\xx_{\text{noise}}$. Following EDM, we use a log-normal distribution for $\sigma$ i.e., $\ln(p(\sigma)) = \cN(P_{\text{mean}}, P_{\text{std}}^2)$ with $P_{\text{mean}} = -1.2$ and $P_{\text{std}}=1.2$.

To train the denoising model, EDM uses preconditioning terms in its objective function to properly scale the inputs and output of the denoiser model $D$. More specifically, the denoising model $D$ is written as 
\begin{align*}
    D(\xx; \ee, \!\sigma)\!:=\!\Big(\frac{\sigma_{\text{data}}}{\sigma^*}\Big)^2\xx + \frac{\sigma\cdot\sigma_{\text{data}}}{\sigma^*} F_\theta\Big(\frac{\xx}{\sigma^*}\,;\ee,\!\frac{\ln(\sigma)}{4}\Big)
\end{align*}
Here, $F_{\theta}$ is a neural network with parameters $\theta$ and $\sigma^* = \sqrt{\sigma^2 + \sigma_{\text{data}}^2}$. We use $\sigma_{\text{data}} = 0.5$.

\paragraph{Sampling} 
Once the denoising model is trained, sampling can be performed by solving the following ODE~\cite{karras2022elucidating}
\begin{align}\label{eq:ode}
    \frac{d\xx}{d\sigma} = -\sigma \nabla_{\xx} \log p(\xx | \ee, \sigma) = \frac{\xx - D(\xx; \ee, \sigma)}{\sigma}
\end{align}
for $\sigma$ flowing backwards from $\sigma = \sigma_{\text{max}}$ to $\sigma=0$. The initial value for $\xx$ is obtained by sampling from the prior distribution $\xx \sim \cN(\bzero, \sigma_{\max}^2 \bI)$. Over the recent years, several samplers have been proposed for sampling from the trained diffusion models~\cite{zhang2023fast, song2021denoising, liu2022pseudo, lu2022dpm, ho2020denoising}. In this paper, we use DEIS~\cite{zhang2023fast} and its stochastic variant~\cite{karras2022elucidating} for synthesizing samples from our model.

\section{Method}\label{sec:method}

Training text-to-video models is much more challenging than training text-to-image diffusion models due to practical difficulties in collecting billion-scale video datasets and securing enough computational resources. Additionally, generating videos is much more challenging since individual frames need to be both photorealistic and temporally coherent. Prior works leverage large-scale image datasets to mitigate these difficulties by either joint training on the image datasets~\cite{wu2022nuwa,ho2022video,ho2022imagen} or finetuning a text-to-image model on the video datasets~\cite{hong2022cogvideo,singer2022make}. Here, we are interested in finetuning text-to-image diffusion models jointly on image and video datasets. We postulate that naively extending the image noise prior to video diffusion is not ideal. We carefully explore the design space of noise priors and propose one that is well suited for our video finetuning task, which leads to significant performance gains.

\setlength{\tabcolsep}{8pt}
\renewcommand{\arraystretch}{1}
\begin{table}[t]
    \centering
    \caption{\new{\textbf{Cosine similarity of the reversed noise.} The noise maps corresponding to the frames sampled from the same videos have a higher similarity than those sampled from different videos.}}
    \label{tab:cosine_sim}
    \tabcspace
    \begin{tabular}{c|c}
    \toprule
     & Cosine Similarity                          \\
    \midrule
    (a) Same video noise      & $0.206$\small{$\pm 0.156$} \\
    (b) Different video noise & $0.001$\small{$\pm 0.009$} \\
    \bottomrule
    \end{tabular}
\end{table}

\begin{figure*}[t!]
    \centering
    \includegraphics[width=\textwidth]{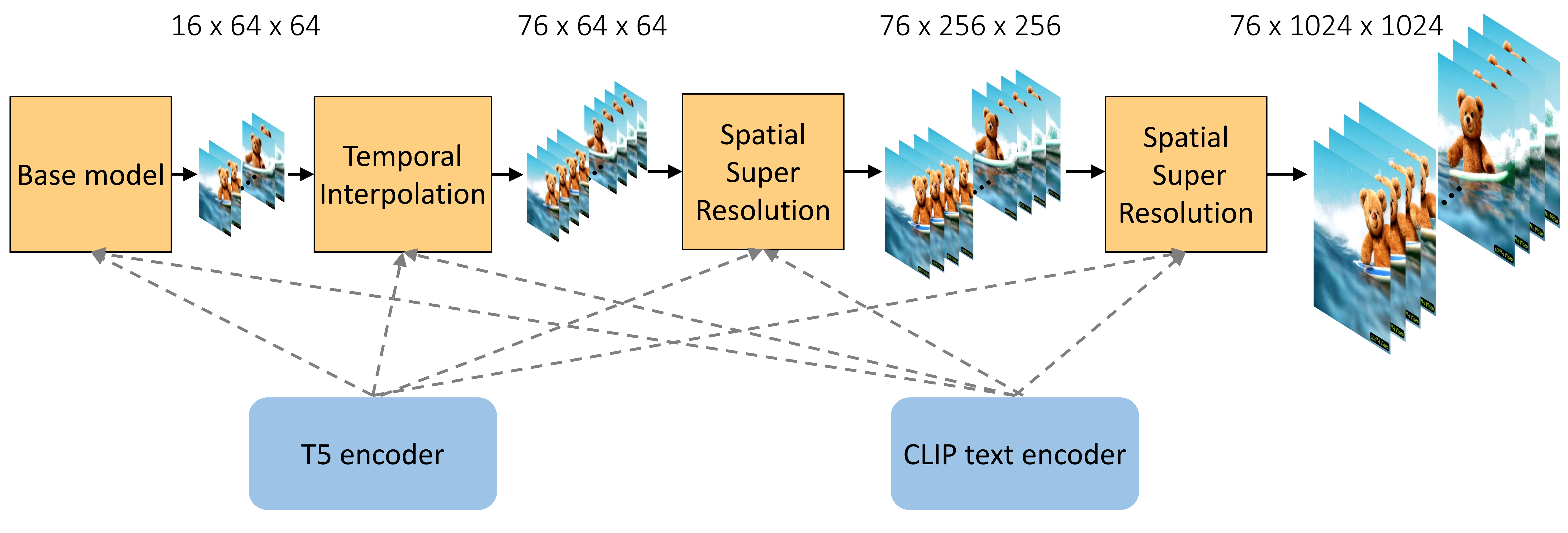}
    \caption{\textbf{Model architecture.} Our pipeline consists of a cascade of four networks --- a base model and three upsampling models. All four models take inputs as the text embeddings obtained from the T5 encoder and the CLIP text encoder. The base model produces $16$ video frames of spatial resolution $64 \times 64$ with a frameskip of $5$. The first upsampling model performs a temporal interpolation, resulting in videos of size $76 \times 64 \times 64$ while the subsequent two super-resolution models perform spatial super-resolution to produce videos of sizes $76 \times 256 \times 256$ and $76 \times 1024 \times 1024$.}
    \label{fig:model_arch}
    \vspace{-0.2cm}
\end{figure*}

\paragraph{Correlated noise model} 
An image diffusion model is trained to denoise independent noise from a perturbed image. The noise vector $\beps$ in the denoising objective \eqref{eq:denoising_objective} is sampled from an i.i.d. Gaussian distribution $\beps \sim \cN(\bzero, \bI)$. However, after training the image diffusion model and applying it to reverse real frames from a video into the noise space in a per-frame manner, we find that the noise maps corresponding to different frames are highly correlated. This is illustrated in Figure~\ref{fig:tsne}, where the t-SNE plot of noise maps corresponding to different video frames are plotted. When the input frames come from the same video (shown in blue dots in Figure~\ref{fig:tsne_a}, noise maps are clustered. The use of i.i.d. sampling (shown in orange dots in Figure~\ref{fig:tsne_b} does not capture these correlations. \new{This is also depicted quantitatively in Table~\ref{tab:cosine_sim} where we compute the average pairwise cosine similarity between noise corresponding to (a) same video and (b) different video. (a) is much higher than (b).} As a result, the video diffusion model \new{trained with i.i.d. noise} is coerced to forget such correlation among the noise between different frames, making it difficult to preserve knowledge from the image diffusion model. Motivated by this observation, we propose to modify the noise process to preserve the correlation between different frames. To this end, we investigate two noising strategies - mixed and progressive noising.

\textbf{Mixed noise model: }
Let $\beps^{1}, \beps^{2}, \hdots \beps^{n_s}$ denote the noise corresponding to individual video frames i.e., $\beps^{i}$ corresponds to the $i^{th}$ element of the noise tensor $\beps$. In the mixed noise model, we generate two noise vectors $\beps_{\text{shared}}$ and $\beps_{\text{ind}}$. $\beps_{\text{shared}}$ is a common noise vector shared among all video frames, while $\beps_{\text{ind}}$ is the individual noise per frame. The linear combination of both these vectors is used as the final noise.
\begin{align}\label{eq:mixed_kaal} 
    \beps_{\text{shared}} \sim \cN\left(\bzero, \frac{\alpha^2}{1 + \alpha^2}\bI \right),& \,   \beps_{\text{ind}}^{i} \sim \cN\left(\bzero, \frac{1}{1 + \alpha^2}\bI \right) \\
    \beps^{i} = \beps_{\text{shared}}& + \beps_{\text{ind}}^{i} \nonumber
\end{align}

\textbf{Progressive noise model: }
In the progressive noise model, the noise for each frame is generated in an autoregressive fashion in which the noise at frame $i$ is generated by perturbing the noise at frame $i-1$. Let $\beps_{\text{ind}}^{i}$ denote the independent noise generated for frame $i$. Then, progressive noising can be formulated as 
\begin{align}\label{eq:progressive_kaal}
    \beps^{0} \sim \cN(\bzero, &\bI) \quad  \beps_{\text{ind}}^{i} \sim \cN(\bzero, \new{\frac{1}{1 + \alpha^2}\bI)} \\
    \beps^{i} &= \new{\frac{\alpha}{\sqrt{1 + \alpha^2}}} \beps^{i-1} + \beps_{\text{ind}}^{i} \nonumber
\end{align}
In both these models, $\alpha$ controls how much noise is shared among different video frames. The higher the value of $\alpha$, the more correlation exists among the noise maps corresponding to different frames. As $\alpha \to \infty$, all frames would have the same noise which results in generating a frozen video. 
On the other hand, $\alpha=0$ corresponds to i.i.d. noise. 

As shown in Figure~\ref{fig:tsne_b}, the use of progressive noise sampling (blue dots) better models the correlations between different noise maps by obtaining similar clustering patterns to the noise maps of real video frames embedded by a pretrained text-to-image model in Figure~\ref{fig:tsne_a} (blue dots).

\paragraph{Model architecture}

As visualized in Figure~\ref{fig:model_arch}, our model consists of a cascade of four networks --- a base network and three upsampling stacks. The base network generates an output video of dimension $16 \times 64 \times 64$ with a frameskip of $5$. It generates the frames $\{ 1, 6, 11, \hdots 76 \}$. The first upsampling network performs a temporal interpolation to produce a video of size $76 \times 64 \times 64$. The second and the third super-resolution network performs spatial upsampling to produce the outputs of sizes $76 \times 256 \times 256$ and $76 \times 1024 \times 1024$. We utilize \ediffi~\cite{balaji2022ediffi}, a state-of-the-art text-to-image diffusion model, to initialize our base and spatial super-resolution models. Similar to prior works~\cite{ho2022video, singer2022make}, we adapt the image-based U-Net model for the video synthesis task by making the following changes: (1) Transforming 2D convolutions to 3D by adding a dimension $1$ to temporal axis and (2) Adding temporal attention layers. Please refer to the supplementary material for more details. 

Similar to Ho \etal~\cite{ho2022video}, we jointly finetune the model on video and image datasets by concatenating videos and images in the temporal axis and applying our temporal modules only on the video part. Similarly to \ediffi, our model uses both T5 text embeddings~\cite{raffel2020exploring} and CLIP text embeddings~\cite{radford2021learning}. We drop each of the embeddings independently at random during training, as in \ediffi.

\section{Experiments}\label{sec:expr}

In this section, we evaluate our proposed strategy of training diffusion models for video synthesis on two sets of experiments. We first comprehensively analyze our proposed noise model on the small-scale UCF-101 dataset. We then scale up our experiments to the challenging large-scale text-to-video synthesis task.



\subsection{Experimental Setups}
We conduct ablation experiments in a small-scale unconditional video generation setting and pick the best configuration for our large-scale text-to-video generation run.

\paragraph{Datasets} We train our model on the UCF-101 dataset~\cite{soomro2012ucf101} for the small-scale experiments, where we follow the protocol defined in Ho~\etal\cite{ho2022video} to generate videos of size $16\times64\times64$. 
\new{UCF-101 dataset contains $13,320$ videos. We randomly sample frames from these videos to train our image synthesis model. For our large-scale experiments, we use a combination of public and proprietary datasets for text-to-image and text-to-video finetuning. Most of the videos are of 2K resolution with 16:9 aspect ratio.} All data was filtered using a preset CLIP and aesthetic scores\footnote{\footnotesize \url{https://github.com/christophschuhmann/improved-aesthetic-predictor}} to ensure high quality. 
Our final image dataset contains around $1.2$ billion text-image pairs and $22.5$ million text-video pairs.

\paragraph{Training details} 
In the unconditional generation experiment on the UCF-101 dataset, to do an ablation study on the model size, we design 3 models where each model has $69$M, $112$M, and $253$M parameters, respectively. As a comparison, 
the baseline Video Diffusion Model (VDM)~\cite{ho2022video} contains $1.2$B parameters. In the large-scale text-to-video experiment, our base and temporal interpolation models contain $1.08$B parameters. Our super-resolution model adapted from the efficient U-Net~\cite{saharia2022photorealistic} architecture with temporal convolution layers~\cite{ho2022imagen,singer2022make} contains $313$M parameters. Please refer to the supplementary material for more training details.

\setlength{\tabcolsep}{0.5pt}
\begin{figure*}[ht!]
    \begin{tabular}{c c c c}
        \includegraphics[width=0.242\textwidth]{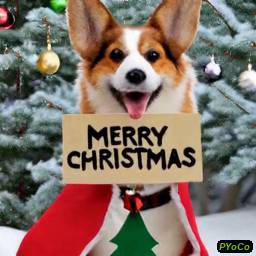}  &
        \includegraphics[width=0.242\textwidth]{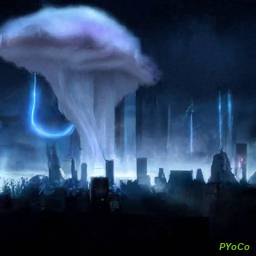}  &
        \includegraphics[width=0.242\textwidth]{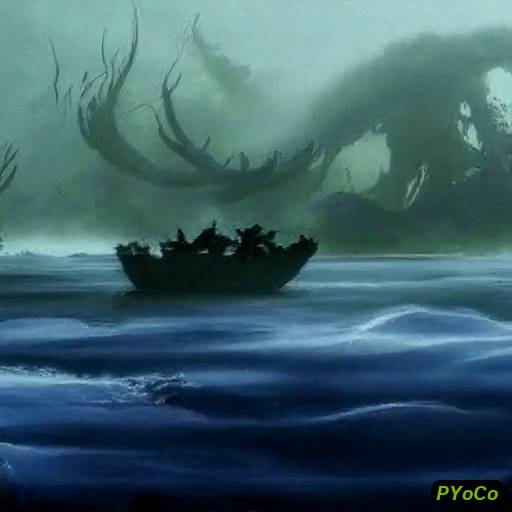}  &
        \includegraphics[width=0.242\textwidth]{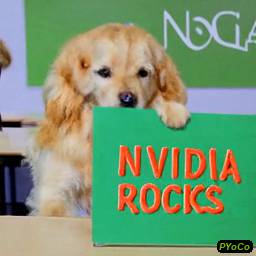} \hspace{0.35em} \\
    \end{tabular}
    \begin{tabularx}{\textwidth}[ht!]{m{0.233\textwidth} c m{0.233\textwidth} c m{0.233\textwidth} c m{0.233\textwidth}}
        \emph{\small A cute corgi wearing a red robe holding a sign that says "Merry Christmas". There is a Christmas tree in the background.} & \hspace{0.55em} &
        \emph{\small An epic tornado attacking above a glowing city at night, the tornado is made of smoke, highly detailed.} & \hspace{0.55em} &
        \emph{\small Small boat sailing in the ocean, giant Cthulhu monster coming out a dense mist in the background, giant waves attacking.} & \hspace{0.55em} &
        \emph{\small A golden retriever puppy holding a green sign that says "NVIDIA ROCKS". Background is a classroom.} \\
    \end{tabularx}
    
    \begin{tabular}{c c c c}
        \includegraphics[width=0.242\textwidth]{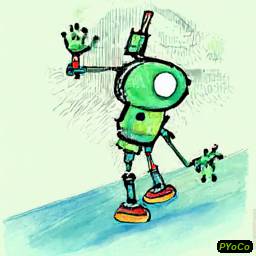}  &
        \includegraphics[width=0.242\textwidth]{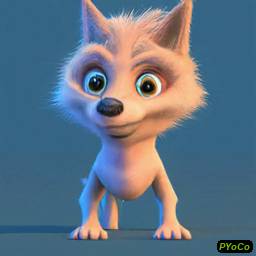}  &
        \includegraphics[width=0.242\textwidth]{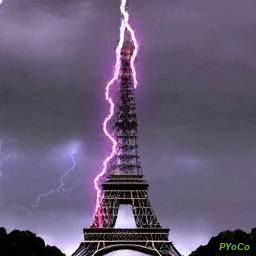}  &
        \includegraphics[width=0.242\textwidth]{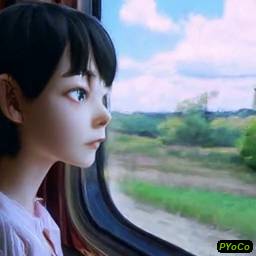} 
        \\
    \end{tabular}
    \begin{tabularx}{\textwidth}[ht!]{m{0.233\textwidth} c m{0.233\textwidth} c m{0.233\textwidth} c m{0.233\textwidth}}
        \emph{\small A cute funny robot dancing, centered, award winning watercolor pen illustration.} & \hspace{0.55em} &
        \emph{\small A cartoon white wolf is giving puppy-dog eyes, detailed fur, very cute kid's film character.} & \hspace{0.55em} &
        \emph{\small A lightning striking atop of eiffel tower, dark clouds in the sky, slow motion.} & \hspace{0.55em} &
        \emph{\small An anime girl looks at the beautiful nature through the window of a moving train, well rendered.} \\
    \end{tabularx}
    
    \begin{tabular}{c c c c}
        \includegraphics[width=0.242\textwidth]{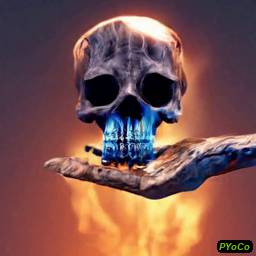}  & 
        \includegraphics[width=0.242\textwidth]{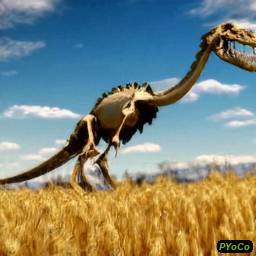}  & 
        \includegraphics[width=0.242\textwidth]{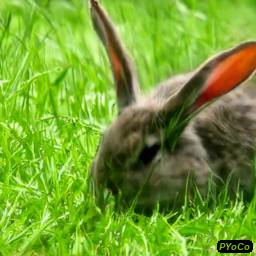} &
        \includegraphics[width=0.242\textwidth]{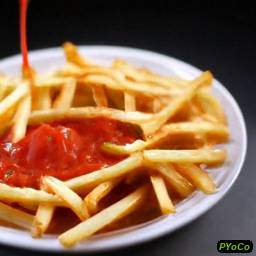}  \\
    \end{tabular}
    \begin{tabularx}{\textwidth}[ht!]{m{0.233\textwidth} c m{0.233\textwidth} c m{0.233\textwidth} c m{0.233\textwidth}}
        \emph{\small A skull burning while being held up by a skeletal hand.} & \hspace{0.55em} &
        \emph{\small A huge dinosaur skeleton walking in a golden wheat field on a bright sunny day.} & \hspace{0.55em} &
        \emph{\small A cute rabbit is eating grass, wildlife photography.} & \hspace{0.55em} &
        \emph{\small Tomato sauce pouring over fries.}
    \end{tabularx}
    
    \caption{Sample generations. \new{\emph{Please check our \href{https://research.nvidia.com/labs/dir/pyoco/}{project website} to view the videos.}}}
    \vspace{-0.2cm}
    \label{fig:visual_generations}
\end{figure*}



\setlength{\tabcolsep}{4pt}
\renewcommand{\arraystretch}{1}

\begin{table}[t!]
\centering
\caption{Zero-shot text to video generation on UCF-101. Our approach gives significant performance gains compared to the prior baselines both in inception score and FVD metrics.}
\label{tab:ucf_zeroshot}
\tabcspace
\begin{tabular}{lcc}
\toprule
Method              & IS  $(\uparrow)$   & FVD  $(\downarrow)$   \\
\midrule
CogVideo~\cite{hong2022cogvideo} (Chinese) & 23.55 & 751.34 \\
CogVideo~\cite{hong2022cogvideo} (English) & 25.27 & 701.59 \\
Make-A-Video~\cite{singer2022make}         & 33.00 & 367.23 \\
MagicVideo~\cite{zhou2022magicvideo}   & - & 655.00 \\
Video LDM~\cite{blattmann2023videoldm}   & 33.45 & 550.61 \\
VideoFactory~\cite{wang2023videofactory}  & - & 410.00 \\
\rowcolor[HTML]{DFDFDF} 
PYoCo       & \textbf{47.76} &  \textbf{355.19} \\
\bottomrule
\end{tabular}
\end{table}

\begin{table}[t!]
\caption{Text conditional zero-shot generation on MSRVTT. Our approach with the base config achieves the best results, and using an ensemble further improves the FIDs.}
\label{tab:msrvtt_zeroshot}
\centering
\begin{tabular}{lcc}
\toprule
Method             & CLIP-FID  $(\downarrow)$   & FID $(\downarrow)$ \\
\midrule
NUWA~\cite{wu2022nuwa} (Chinese)     & 47.68 & - \\
CogVideo~\cite{hong2022cogvideo} (Chinese) & 24.78 & - \\
CogVideo~\cite{hong2022cogvideo} (English) & 23.59 & - \\
Make-A-Video~\cite{singer2022make}  & 13.17 & - \\
MagicVideo~\cite{zhou2022magicvideo} & - & 36.50 \\
Latent-Shift~\cite{an2023latent} & 15.23 & - \\
\rowcolor[HTML]{DFDFDF} PYoCo (Config-A) & 10.21 & 25.39 \\
\rowcolor[HTML]{DFDFDF} PYoCo (Config-B) & 9.95  & 24.28 \\
\rowcolor[HTML]{DFDFDF} PYoCo (Config-C) & 9.91  & 24.54 \\
\rowcolor[HTML]{DFDFDF} PYoCo (Config-D) & \textbf{9.73}  & \textbf{22.14} \\
\bottomrule
\end{tabular}
\end{table}

\paragraph{Evaluation} For the small-scale experiments on UCF-101 dataset, we follow the protocol defined in the prior approaches~\cite{tian2021a,skorokhodov2021stylegan,ho2022video} and report the Inception Score (IS)~\cite{saito2020train} calculated by a trained C3D model~\cite{tran2015learning} and Fréchet Video Distance (FVD)~\cite{unterthiner2018towards} by a trained I3D model~\cite{carreira2017quo}. For the large-scale text-to-video experiments, we perform the zero-shot evaluation of the video generation quality on the UCF-101 and MSR-VTT datasets following Make-A-Video~\cite{singer2022make}. We carefully discuss the evaluation process below.

\paragraph{UCF-101 experiment} We use IS and FVD for evaluation in our small-scale experiments. UCF-101 is a categorical video dataset designed for action recognition. When sampling from the text-to-video model, we devise a set of prompts for each class name to be used as the conditional input. This is necessary as some class names (such as \textit{jump rope}) are not descriptive. We list all the prompts we use in the supplementary material. We sample $20$ videos for each prompt to compute the IS metric. For FVD, we follow the prior work~\cite{le2021ccvs,tian2021a} and sample $2,048$ videos for evaluation.

\paragraph{MSR-VTT experiment} MSR-VTT~\cite{xu2016msr} test set contains $2,990$ videos as well as $59,794$ captions. All the videos have the same resolution of $320\times240$. We generate a $76\times256\times256$ video for each $59,794$ caption and save the videos in an \textit{mp4} format with a high bit rate. To compare with Make-A-Video, we compute FID using a ViT-B/32 model~\cite{kynkaanniemi2023the}. We also report a more common FID metric computed by an Inception-V3 model. We also examine the idea of ensemble denoiser~\cite{balaji2022ediffi} by finetuning the level-1 experts of each model. We denote Config-A as the configuration of using only baseline models and Config-B to Config-D as incrementally changing super-resolution model, temporal interpolation model, and base model with the corresponding ensemble models.

\subsection{Main Results}

\begin{figure*}[ht!]
    \begin{tabularx}{\textwidth}[ht!]{c m{0.2\textwidth} c m{0.2\textwidth} c m{0.2\textwidth} c m{0.2\textwidth}}
        \hspace{1.4em} & Make-A-Video~\cite{singer2022make} & \hspace{2.35em} & \ediffivideo & \hspace{3.5em} & Imagen Video~\cite{ho2022imagen} & \hspace{4.9em} & \ediffivideo \\
    \end{tabularx}
    \begin{tabular}{c c c c c}
        \includegraphics[height=0.208\textwidth]{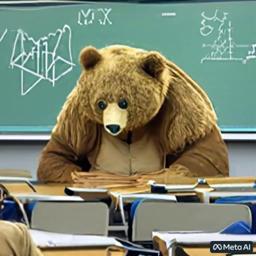} &
        \includegraphics[height=0.208\textwidth]{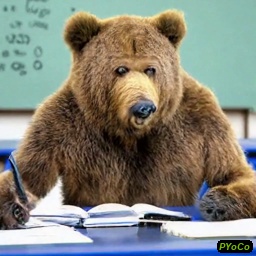} & \hspace{0.3em} &
        \includegraphics[height=0.208\textwidth]{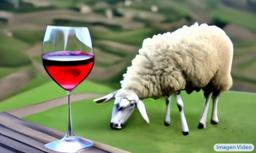} &
        \includegraphics[height=0.208\textwidth]{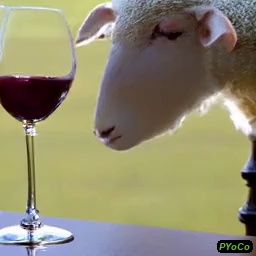} \\
    \end{tabular}

    \begin{tabularx}{\textwidth}[ht!]{c m{0.37\textwidth} c m{0.37\textwidth}}
        \hspace{3.5em} & \emph{\small A confused grizzly bear in calculus class.} & \hspace{7.0em} &
        \emph{\small A sheep to the right of a wineglass.}  \\
    \end{tabularx}
    \begin{tabular}{c c c c c}
        \includegraphics[height=0.208\textwidth]{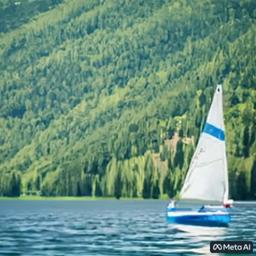} &
        \includegraphics[height=0.208\textwidth]{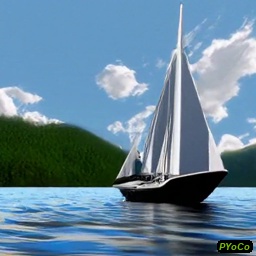} & \hspace{0.3em} &
        \includegraphics[height=0.208\textwidth]{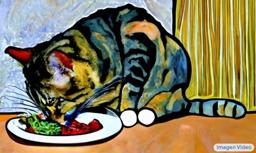}  &
        \includegraphics[height=0.208\textwidth]{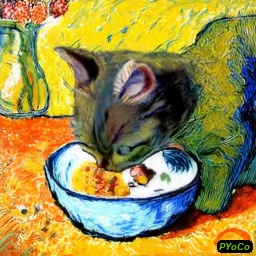} \\
    \end{tabular}
    \begin{tabularx}{\textwidth}[ht!]{c m{0.37\textwidth} c m{0.37\textwidth}}
        \hspace{2.1em} & \emph{\small Sailboat sailing on a sunny day in a mountainlake.} & \hspace{6.0em} &
        \emph{\small A cat eating food out of a bowl, in style of Van Gogh.}  \\
    \end{tabularx}
\captionof{figure}{Qualitative comparison with baseline approaches. The two panels on the left show the comparison of our approach with Make-A-Video~\cite{singer2022make}, while those on the right show the comparison with Imagen Video~\cite{ho2022imagen}. \ediffivideo achieves better photorealism compared to the two approaches.}
\label{fig:qualitative_comparison}
\end{figure*}

\paragraph{Large-scale text-to-video synthesis}
We quantitatively compare our method against Make-A-Video~\cite{singer2022make}, NUWA~\cite{wu2022nuwa}, CogVideo~\cite{hong2022cogvideo}, and several concurrent works~\cite{blattmann2023videoldm,zhou2022magicvideo,blattmann2023videoldm,wang2023videofactory,an2023latent}. Table~\ref{tab:ucf_zeroshot} shows that our method outperforms all the baselines on the UCF-101 dataset and improves the zero-shot Inception Score from $33.45$ to $47.76$.
In Table~\ref{tab:msrvtt_zeroshot}, we show that our baseline model achieves a new state-of-the-art CLIP-FID score~\cite{kynkaanniemi2023the} of $10.21$, while using ensemble models further improves both CLIP-FID and FID scores.
In Figure~\ref{fig:visual_generations}, we qualitatively visualize the synthesis capability of our approach. Our model achieves high-quality zero-shot video synthesis capability with good photorealism and temporal coherency. We also provide a qualitative comparison with Make-A-Video~\cite{singer2022make} and Imagen Video~\cite{ho2022imagen} in Figure~\ref{fig:qualitative_comparison}. We observe that our model is able to produce videos with better details than both approaches, as shown in the animal videos. We also produce better-stylized videos than Imagen Video.

\paragraph{Small-scale unconditional video synthesis}

\setlength{\tabcolsep}{8pt}
\renewcommand{\arraystretch}{1}

\begin{table}[t]
\centering
\caption{Unconditional UCF-101 generation results. Our approach achieves the state-of-the-art inception score and FVD, while having considerably smaller parameter count compared to other diffusion-based approaches such as VDM (1B parameters).}
\tabcspace
\label{tab:main_small_scale}
\begin{tabular}{lcc}
\toprule
Method & IS $(\uparrow)$  & FVD  $(\downarrow)$  \\
\midrule
TGAN~\cite{saito2017temporal} & 15.83\tiny{$\pm .18$} & -  \\
LDVD-GAN~\cite{kahembwe2020lower} & 22.91\tiny{$\pm .19$}   & - \\
VideoGPT~\cite{yan2021videogpt} & 24.69\tiny{$\pm .30$}   & - \\
MoCoGAN-HD~\cite{tian2021a}  & 32.36\phantom{\tiny{$\pm .00$}} & 838 \\
DIGAN~\cite{yu2021generating} & 29.71\tiny{$\pm .53$}   & 655\tiny{$\pm 22$}   \\
CCVS~\cite{le2021ccvs}  & 24.47\tiny{$\pm .13$} & 386\tiny{$\pm 15$} \\
StyleGAN-V~\cite{skorokhodov2021stylegan}  & 23.94\tiny{$\pm .73$} & - \\
VDM~\cite{ho2022video}  & 57.00\tiny{$\pm .62$} & - \\
TATS~\cite{ge2022long}  & 57.63\tiny{$\pm .73$} & 430 \tiny{$\pm 18$} \\
\rowcolor[HTML]{DFDFDF} 
\ediffivideo ($112$M) & 57.93\tiny{$\pm .24$} & 332 \tiny{$\pm 13$} \\
\rowcolor[HTML]{DFDFDF} 
\ediffivideo ($253$M) & \textbf{60.01}\tiny{$\pm .51$} & \textbf{310} \tiny{$\pm 13$} \\
\bottomrule
\end{tabular}
\tabspace
\end{table}

We report IS and FVD scores on UCF-101 dataset in Table~\ref{tab:main_small_scale} and compare our model with multiple unconditional video generation baselines. Note that using class labels as conditional information could lead to sizeable improvement in IS and FVD scores~\cite{ge2022long}, which we do not consider as the comparison. Our method attains state-of-the-art unconditional video generation quality. Compared with previous diffusion-based unconditional generation model~\cite{ho2022video}, our model is $\sim10\times$ smaller and has $\sim14\times$ less training time ($75$ GPU-days vs. $925$ GPU-days).

\vspace{-2mm}
\setlength{\tabcolsep}{0.5pt}

\vspace{2mm}

\setlength{\tabcolsep}{6pt}
\renewcommand{\arraystretch}{1}

\begin{table}[t]
\centering
\caption{Quantitative results of different training strategies \new{on UCF-101 dataset.}}
\tabcspace
\label{tab:strategies}
\begin{tabular}{l|lll}
\toprule
                 & IS$(\uparrow)$ & FVD $(\downarrow)$ &  FID $(\downarrow)$ \\
\midrule
Image Diffusion (ID)  & -     & -      & 30.05  \\ \hline
Training from scratch  & 28.25 & 903.37 & 124.75 \\
Finetuning from ID      & 41.25 & 566.67 & 56.43  \\
\rowcolor[HTML]{DFDFDF} 
+ Mixed Noise        & 52.71 & \textbf{337.40} & \textbf{31.57}  \\
\rowcolor[HTML]{DFDFDF} 
+ Progressive Noise  & \textbf{53.52} & 339.67 & 31.88  \\
\bottomrule
\end{tabular}
\tabspace
\end{table}
\subsection{Ablation Study}
We quantitatively compare several training strategies for video diffusion models. Then, we perform ablation on the correlation ratio in the Equations \ref{eq:mixed_kaal} and \ref{eq:progressive_kaal}, a key hyper-parameter in our approach.

\paragraph{Training strategies} We compare training from scratch, a simple finetuning baseline, finetuning with mixed noising, and progressive noising using IS, FVD, and averaged frame FID metrics \new{on the UCF-101 dataset} in Table~\ref{tab:strategies}. We first find that finetuning from an image diffusion model is much more effective than training from scratch. For finetuning from the image model, the correlated noise model produces better video generation quality than the independent noise model. In addition, we notice that the correlated noise better preserves the image quality learned by the pretrained image model and produces a lower frame FID. This is particularly desired in large-scale text-to-video training to fulfill the goal of inheriting the knowledge from the image model missing in the video datasets. \new{Specifically, most videos contain realistic scenes captured by cameras and have infrequent media types like paintings, illustrations, sketches, etc. 
Moreover, the video data is much smaller in volume, and the scenes are less diverse than image datasets.} As shown in Figure~\ref{fig:visual_generations}, our model can preserve properties learned from image datasets that are not presented in our video dataset, such as the artistic styles, and generate faithful motion on them.

\begin{figure*}[t!]
    \setlength{\tabcolsep}{0.5pt}
    \renewcommand{\arraystretch}{0.5}
    \begin{tabular}{c c c}
        \includegraphics[width=0.325\linewidth,trim=0 0 0 0,clip]{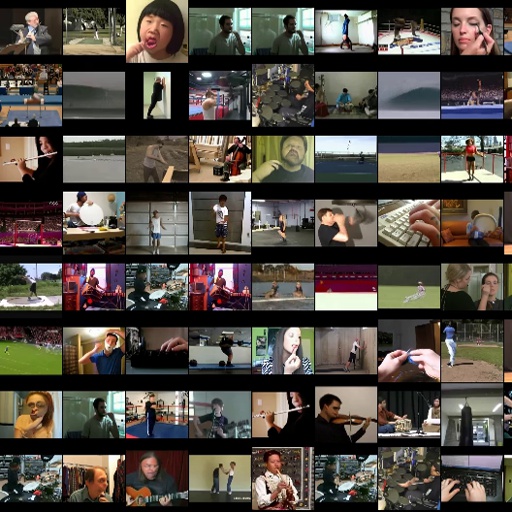} &
        \includegraphics[width=0.325\linewidth,trim=0 0 0 0,clip]{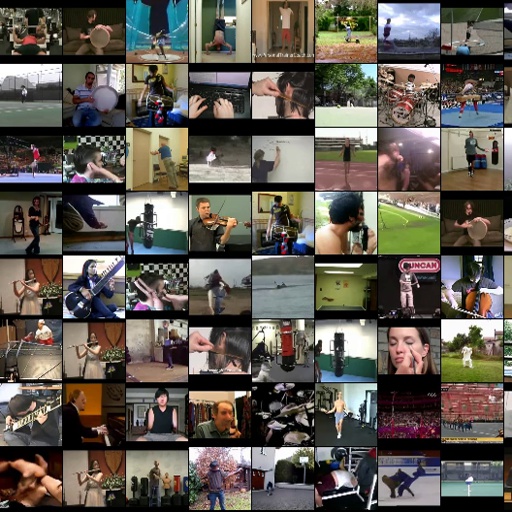} &
        \includegraphics[width=0.325\linewidth,trim=0 0 0 0,clip]{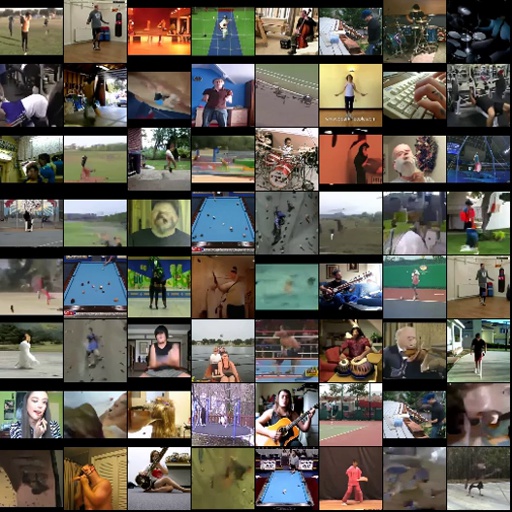}
    \end{tabular}
    \begin{tabularx}{\linewidth}{c m{0.3\linewidth} c m{0.3\linewidth} c m{0.318\linewidth}} 
        \hspace{7.3em} & $\alpha=0$ & \hspace{0.7em} &
        $\alpha=1$ & \hspace{0.7em} & 
        $\alpha=10$
    \end{tabularx}
    \caption{\textbf{Visual ablation on $\alpha$.} Small $\alpha=0$ reduces video quality and diversity and large $\alpha=10$ yields motion artifacts. }
    \label{fig:alpha_vis}
\end{figure*}
\newcommand{\MixedKaalFVD}{
(0, 563.32666667)
(1, 526.96)
(2, 525.22)
(3, 361.6)
(4, 327.4)
(5, 381.59333333)
(6, 544.97)
(7, 562.27333333)
(8, 1165.62666667)
}

\newcommand{\ProgKaalFVD}{
(0, 563.32666667)
(1, 547.50666667)
(2, 471.77)
(3, 443.34666667)
(4, 376.8)
(5, 322.59)
(6, 425.40666667)
(7, 453.61666667)
(8, 1165.62666667)
}

\newcommand{\KaalFVD}{
(0, 903.37)
(1, 903.37)
(2, 903.37)
(3, 903.37)
(4, 903.37)
(5, 903.37)
(6, 903.37)
(7, 903.37)
(8, 903.37)
}

\newcommand{\MixedKaalIS}{
(0, 42.11666667)
(1, 43.05333333)
(2, 43.80666667)
(3, 49.67)
(4, 52.71)
(5, 49.01666667)
(6, 37.77)
(7, 45.09666667)
(8, 31.43666667)
}

\newcommand{\ProgKaalIS}{
(0, 42.11666667)
(1, 42.32666667)
(2, 43.94)
(3, 45.23)
(4, 50.44)
(5, 53.52)
(6, 49.21)
(7, 46.45666667)
(8, 31.43666667)
}

\newcommand{\KaalIS}{
(0, 28.25)
(1, 28.25)
(2, 28.25)
(3, 28.25)
(4, 28.25)
(5, 28.25)
(6, 28.25)
(7, 28.25)
(8, 28.25)
}

\vspace{-2mm}
\renewcommand{\plothh}{90mm}
\renewcommand{\plothc}{85mm}
\renewcommand{\plotvv}{50mm}
\begin{figure}[t!]
    \centering
    \scriptsize
    \begin{tikzpicture}
        \begin{axis}[
                width=\plothh, height=\plotvv,
                xmin={-0.3}, xmax={8.3}, xmode={linear},   
                xtick={0, 1, 2, 3, 4, 5, 6, 7, 8},
                xticklabels={$0$, $0.1$, $0.2$, $0.5$, $1.0$, $2.0$, $5.0$, $10.0$, $\infty$},
                xlabel={$\alpha$}, xlabel near ticks,
                ymin={250}, ymax={1300}, ymode={linear},
                ytick={500, 1000, 1300},
                yticklabels={$500$, $1000$, \tickFVD{}},
                grid={major},
                legend pos={north west},
                legend cell align={left},
            ]
            \addplot[C0] coordinates {\MixedKaalFVD};
            \addplot[C1] coordinates {\ProgKaalFVD};
            \addplot[black, dashed] coordinates {\KaalFVD};
            \addplot[C0, only marks, forget plot] coordinates {\MixedKaalFVD};
            \addplot[C1, only marks, forget plot] coordinates {\ProgKaalFVD};
            \legend{
                {Mixed},
                {Progressive},
                {Scratch},
            }
        \end{axis}
    \end{tikzpicture}
    \caption{\textbf{Quantitative ablation on hyperparameter $\alpha$.} Finetuning with temporally correlated prior improves over training from scratch. Using a too-large or too-small $\alpha$ leads to inferior results. $\alpha=1$, $\alpha=2$ each works the best for mixed and progressive noising, respectively.}
    \label{fig:sweep_alpha}
\end{figure}
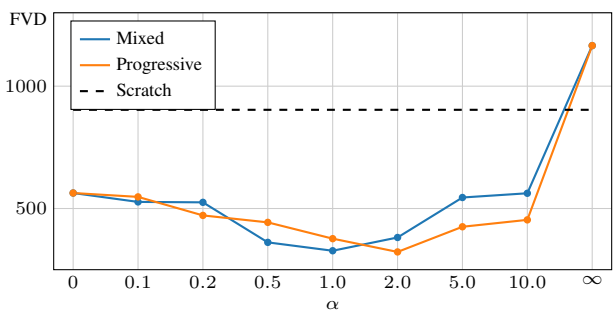

\paragraph{Correlation ratio} The hyperparameter $\alpha$ in the Equations \ref{eq:mixed_kaal} and \ref{eq:progressive_kaal} controls the correlation between the noise of different frames. A larger $\alpha$ injects more correlation into the noise. The correlation disappears when $\alpha \rightarrow 0$, and the mixed and progressive noise models reproduce the vanilla noise model. 
To find optimal $\alpha$, we train our UCF-small model ($69$M parameters) using $\alpha \in \{0, 0.1, 0.2, 0.5, 1, 1, 2, 5, 10, \infty\}$ and report FVD in Figure~\ref{fig:sweep_alpha}. 
For each $\alpha$ value, we repeat the experiment 3 times and report the mean. Note that $\alpha=0$ indicates finetuning with the independent frame noise, and $\alpha=\infty$ indicates using identical noise maps for all the frames, which produces frozen videos during the inference time. Finetuning an image diffusion model almost consistently outperforms the training-from-scratch baseline with different $\alpha$s. Using $\alpha=1$ for mixed noising and $\alpha=2$ for progressive noising produces similar best results. \new{We also show qualitiative results for models trained with $\alpha=0, 1, 10$ in Figure~\ref{fig:alpha_vis}. When $\alpha$ is too small, we notice a degradation in visual quality in the generated video frames and a reduced video diversity. For example, we notice many repeated samples and black borders in almost every video generated with $\alpha=0$. On the other hand, when $\alpha$ is too large, the model has difficulty generating proper motions.}

\paragraph{Model size} 
We pick the best $\alpha$ for the mixed and progressive noise models and compare them with the model trained from scratch on models with different numbers of parameters, $69$M, $112$M, and $253$M.
Figure~\ref{fig:sweep_modelsize} shows that our mixed and progressive models outperform the baseline consistently by a large margin in terms of FVD.
Overall, mixed and progressive noising provide similar performance. 
In our large large-scale experiments, we choose progressive noising with $\alpha=2$ due to its autoregressive nature.

\section{Conclusion}\label{sec:conclusion}

We proposed a new efficient way of training text-to-video generation models.
By observing that the noise maps generating the frames of a video are clustered together, we study mixed and progressive noise priors well-suited for sequential video frame generation. 
We apply our progressive noise prior to finetuning a state-of-the-art diffusion-based text-to-image model to achieve a state-of-the-art large-scale text-to-video model. The high quality of the generated videos and the state-of-the-art Inception and FID scores demonstrate the strength of our approach.

\newcommand{\ModelEffMixedKaalFVD}{
(68, 337.4)
(112, 322.57)
(253, 310.13)
}

\newcommand{\ModelEffProgKaalFVD}{
(68, 339.67)
(112, 322.59)
(253, 298.88)
}

\newcommand{\ModelEffKaalFVD}{
(68, 903.37)
(112, 748.67)
(253, 642.75)
}

\newcommand{\ModelEffMixedKaalIS}{
(68, 53.52)
(112, 57.93)
(253, 60.01)
}

\newcommand{\ModelEffProgKaalIS}{
(68, 53.52)
(112, 57.31)
(253, 59.82)
}

\newcommand{\ModelEffKaalIS}{
(68, 28.25)
(112, 31.16)
(253, 38.28)
}

\vspace{-2mm}
\renewcommand{\plothh}{90mm}
\renewcommand{\plothc}{85mm}
\renewcommand{\plotvv}{50mm}
\begin{figure}[t!]
    \centering
    \small
    \scriptsize
    \begin{tikzpicture}
        \begin{axis}[
                width=\plothh, height=\plotvv,
                ytick={400, 600, 800, 950},
                yticklabels={$400$, $600$, $800$, \tickFVD{}},
                xlabel={Number of Parameters (M)}, xlabel near ticks,
                grid={major},
                legend pos={north east},
                legend cell align={left},
            ]
            \addplot[C0] coordinates {\ModelEffMixedKaalFVD};
            \addplot[C1] coordinates {\ModelEffProgKaalFVD};
            \addplot[black, dashed] coordinates {\ModelEffKaalFVD};
            \addplot[C0, only marks, forget plot] coordinates {\ModelEffMixedKaalFVD};
            \addplot[C1, only marks, forget plot] coordinates {\ModelEffProgKaalFVD};
            \legend{
                {Mixed ($\alpha=1$)},
                {Progressive ($\alpha=2$)},
                {Scratch},
            }
        \end{axis}
    \end{tikzpicture}
    \caption{\textbf{Ablation on model size.} Larger models consistently improve the performance of both finetuning and training from scratch. Finetuning from image model consistently outperforms training from scratch.}
    \label{fig:sweep_modelsize}
\end{figure}
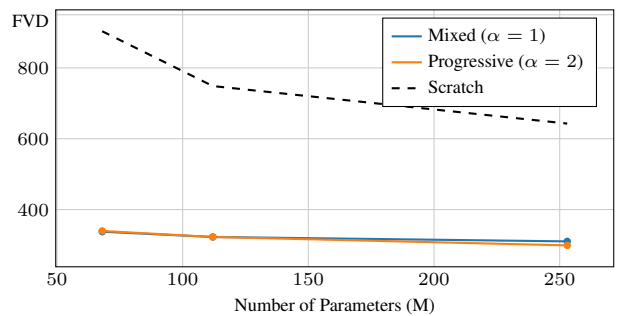

\topic{Acknowledgment.} We would like to thank Amanda Moran, John Dickinson, Sivakumar Arayandi Thottakara, David Page, Ranjitha Prasanna, Venkata Karri, and others in NGC and PBSS team for the computing infrastructure support. We also give thanks to Qinsheng Zhang, Zekun Hao, Tsung-Yi Lin, Ajay Jain, and Chen-Hsuan Lin for useful discussions and feedback. Thanks also go to Yin Xi and Thomas Hayes for clarifying the evaluation protocol of Make-A-Video and to Ming Ding for the details of CogVideo.  
\new{We thank the ICCV PCs, ACs, and reviewers for their service and valuable feedback.}
This work is partly supported by NSF grants No. IIS-1910132 and IIS-2213335.

\clearpage
\appendix

\renewcommand{\thetable}{\Alph{table}}
\renewcommand{\thefigure}{\Alph{figure}}
\renewcommand{\theequation}{\Alph{equation}}

\section{Experimental Setups}
\label{sec:addtional_details}

In this section, we provide additional details of our experiments in terms of implementation, dataset, evaluation, model, and training.

\subsection{Implementation details}

Similar to prior works~\cite{ho2022video, singer2022make}, we adapt the image-based U-Net model for the video synthesis task by making the following changes: (1) We transform the 2D convolution layers to 3D by adding a dimension of $1$ to the temporal axis. For instance, we convert a $3 \times 3$ convolution layer to $1 \times 3 \times 3$ layer. (2) We replace the attention layers in the base and temporal interpolation models with a cascade of spatial and temporal attention layers. The spatial attention layers are reused from \ediffi~\cite{balaji2022ediffi}, while the temporal attention layers are initialized randomly with a projection layer at the end using zero-initialization. We apply temporal attention to the activation maps obtained by moving the spatial dimension of the feature tensor to the batch axis. (3) For the temporal interpolation model, we concatenate the input noise in the channel axis with $16$ frames by infilling $4$ real frames with zero frames. (4) We add a $3 \times 1 \times 1$ convolution layer at the end of each efficient block of the super-resolution model~\cite{saharia2022photorealistic}. (5) For all the models, we apply spatial attention to the reshaped activation maps obtained by moving the temporal dimension of the feature tensor to the batch axis. We apply the same operation to the feature maps input the GroupNorm~\cite{wu2018group} to mimic better the statistics the image model learned. We use cross-attention layers (between text and videos) only in the spatial attention block, as adding it to the temporal attention resulted in significant memory overhead. (6)  We utilize \ediffi~\cite{balaji2022ediffi} to initialize our base and spatial super-resolution models. We use a similar model architecture as the base model for our temporal interpolation model, as they share the same function of hallucinating unseen frames. After finetuning the base model for some time, we use its checkpoint to initialize the temporal interpolation model. (7) Similar to Ho \etal~\cite{ho2022video}, we jointly finetune the model on video and image datasets by concatenating videos and images in the temporal axis and applying our temporal modules only on the video part. (8) Similarly to \ediffi, our model uses both T5~\cite{raffel2020exploring} text embeddings and CLIP text embeddings~\cite{radford2021learning}. During training, we drop each of the embeddings independently at random, as in \ediffi.

\subsection{Dataset and evaluation details}
\label{sec:data_eval_details}

\paragraph{Caption templates for categorical video datasets} Given the name of the category [\textit{class}] such as \textit{kayaking} and \textit{yoga}, we consider the following templates to create video captions:

\begin{itemize}
    \item a man is [\textit{class}].
    \item a woman is [\textit{class}].
    \item a kid is [\textit{class}].
    \item a group of people are [\textit{class}].
    \item doing [\textit{class}].
    \item a man is doing [\textit{class}].
    \item a woman is doing [\textit{class}].
    \item a kid is doing [\textit{class}].
    \item a group of people are doing [\textit{class}].
    \item ~[\textit{class}].
\end{itemize}

\paragraph{Prompts used for UCF-101 evaluation} In our initial explorations, we find that the original class labels in the UCF-101 dataset often cannot describe the video content correctly. For example, the class \textit{jump rope} is more likely describing an object rather than a complete video. Therefore, we write one sentence for each class as the caption for video generation. We list these prompts for evaluating text-to-video generation models on the standard UCF-101 benchmark below~\footnote{A copy-paste friendly version is available in the Google Spreadsheet at \url{https://docs.google.com/spreadsheets/d/1teEGth-Iy1be4Tx7xfXUKBA3aGZ9Hhr2gueTpuuwv94/edit?usp=sharing}}.

\textit{applying eye makeup, applying lipstick, archery, baby crawling, gymnast performing on a balance beam, band marching, baseball pitcher throwing baseball, a basketball player shooting basketball, dunking basketball in a basketball match, bench press, biking, billiards, blow dry hair, blowing candles, body weight squats, a person bowling on bowling alley, boxing punching bag, boxing speed bag, swimmer doing breast stroke, brushing teeth, clean and jerk, cliff diving, bowling in cricket gameplay, batting in cricket gameplay, cutting in kitchen, diver diving into a swimming pool from a springboard, drumming, two fencers have fencing match indoors, field hockey match, gymnast performing on the floor, group of people playing frisbee on the playground, swimmer doing front crawl, golfer swings and strikes the ball, haircuting, a person hammering a nail, an athlete performing the hammer throw, an athlete doing handstand push up, an athlete doing handstand walking, massagist doing head massage to man, an athlete doing high jump, group of people racing horse, person riding a horse, a woman doing hula hoop, man and woman dancing on the ice, athlete practicing javelin throw, a person juggling with balls, a young person doing jumping jacks, a person skipping with jump rope, a person kayaking in rapid water, knitting, an athlete doing long jump, a person doing lunges with barbell, military parade, mixing in the kitchen, mopping floor, a person practicing nunchuck, gymnast performing on parallel bars, a person tossing pizza dough, a musician playing the cello in a room, a musician playing the daf, a musician playing the indian dhol, a musician playing the flute, a musician playing the guitar, a musician playing the piano, a musician playing the sitar, a musician playing the tabla, a musician playing the violin, an athlete jumps over the bar, gymnast performing pommel horse exercise, a person doing pull ups on bar, boxing match, push ups, group of people rafting on fast moving river, rock climbing indoor, rope climbing, several people rowing a boat on the river, couple salsa dancing, young man shaving beard with razor, an athlete practicing shot put throw, a teenager skateboarding, skier skiing down, jet ski on the water, sky diving, soccer player juggling football, soccer player doing penalty kick in a soccer match, gymnast performing on still rings, sumo wrestling, surfing, kids swing at the park, a person playing table tennis, a person doing TaiChi, a person playing tennis, an athlete practicing discus throw, trampoline jumping, typing on computer keyboard, a gymnast performing on the uneven bars, people playing volleyball, walking with dog, a person standing and doing pushups on the wall, a person writing on the blackboard, a kid playing Yo-Yo}

\subsection{Training details}
\label{sec:training_details}
\paragraph{UCF-101 experiments.} For image pretraining phase on the UCF-101 frames, we use an ADAM optimizer with a base learning rate of $2e-4$. For video finetuning phase, we adopt an ADAM optimizer with a base learning rate of $1e-4$. We use a linear warm up of $5,000$ steps for both phases. For sampling, we use stochastic DEIS sampler~\cite{zhang2022fast, karras2022elucidating} with 3kutta, order $6$ and $25$ steps.

\paragraph{Large-scale experiments.} The hyper-parameters we use for the large-scale text-to-video experiments are provided in Table~\ref{tab:hyperparams}.


\begin{table}[ht!]
    \centering
    \caption{Hyperparameters}
    \vspace{-1mm}
    \label{tab:hyperparams}
    \begin{tabular}{l c}
        \toprule
        \multicolumn{2}{c}{Hyperparameters for large-scale experiments}\\
        \midrule
        Optimizer & AdamW \\
        Learning rate & $0.0001$ \\
        Weight decay & $0.01$ \\
        Betas & (0.9, 0.999) \\
        EMA & $0.9999$ \\
        CLIP text embedding & \multirow{2}{*}{$0.2$} \\
        dropout rate & \\
        T5 text embedding  & \multirow{2}{*}{$0.25$} \\
        dropout rate & \\
        Gradient checkpointing & Enabled \\
        \# iterations for base model & $150K$ \\
        \# iterations for super-res model & $220K$ \\
        \multirow{2}{*}{Sampler for base model} & Stochastic DEIS~\cite{zhang2023fast, karras2022elucidating},  \\
        & 3kutta, Order 3, 60 steps \\
        \multirow{2}{*}{Sampler for super-res models} & DEIS, 3kutta \\
        & Order 3, 20 steps \\
        \bottomrule
    \end{tabular}
\end{table}

\subsection{Architecture details}
The architectures used for the small-scale UCF experiments are provided in Tables~\ref{tab:ucf_arch_small}, \ref{tab:ucf_arch_medium} and \ref{tab:ucf_arch_large}. For the large-scale experiment, the architectures used for base model, temporal interpolation model, and the two spatial super-resolution stacks are provided in tables~\ref{tab:text2video_arch_base}, \ref{tab:text2video_arch_ti}, \ref{tab:text2video_arch_sr256} and \ref{tab:text2video_arch_sr1024} respectively.

\label{sec:architecture_details}

\renewcommand{\arraystretch}{1}

\begin{table}[!ht]
    \centering
    \caption{Small (69M parameters) UCF-101 model architecture.}
    \vspace{-1mm}
    \label{tab:ucf_arch_small}
    \begin{tabular}{l c}
        \toprule
        \multicolumn{2}{c}{Small (69M parameters) UCF-101 model}\\
        \midrule
        Channel multiplier & $[1, 2, 2, 3]$ \\
        Dropout & $0.1$ \\
        Number of channels & $128$ \\
        Number of residual blocks & $2$ \\
        Spatial self attention resolutions & $[32, 16, 8]$ \\
        Spatial cross attention resolutions & $[32, 16, 8]$ \\
        Temporal attention resolution & $[32, 16, 8]$ \\
        Number of channels in attention heads & 64\\
        Use scale shift norm & True \\
        \bottomrule
    \end{tabular}
\end{table}

\begin{table}[!ht]
    \centering
    \caption{Medium (112M parameters) UCF-101 model architecture.}
    \vspace{-1mm}
    \label{tab:ucf_arch_medium}
    \begin{tabular}{l c}
        \toprule
        \multicolumn{2}{c}{Medium (112M parameters) UCF-101 model}\\
        \midrule
        Channel multiplier & $[1, 2, 3, 4]$ \\
        Dropout & $0.1$ \\
        Number of channels & $128$ \\
        Number of residual blocks & $2$ \\
        Spatial self attention resolutions & $[32, 16, 8]$ \\
        Spatial cross attention resolutions & $[32, 16, 8]$ \\
        Temporal attention resolution & $[32, 16, 8]$ \\
        Number of channels in attention heads & 64\\
        Use scale shift norm & True \\
        \bottomrule
    \end{tabular}
\end{table}

\begin{table}[!ht]
    \centering
    \caption{Large (253M parameters) UCF-101 model architecture.}
    \vspace{-1mm}
    \label{tab:ucf_arch_large}
    \begin{tabular}{l c}
        \toprule
        \multicolumn{2}{c}{Large (253M parameters) UCF-101 model}\\
        \midrule
        Channel multiplier & $[1, 2, 3, 4]$ \\
        Dropout & $0.1$ \\
        Number of channels & $192$ \\
        Number of residual blocks & $2$ \\
        Spatial self attention resolutions & $[32, 16, 8]$ \\
        Spatial cross attention resolutions & $[32, 16, 8]$ \\
        Temporal attention resolution & $[32, 16, 8]$ \\
        Number of channels in attention heads & 64\\
        Use scale shift norm & True \\
        \bottomrule
    \end{tabular}
\end{table}

\vfill\null

\begin{table}[!ht]
    \centering
    \caption{Architecture for the base model in text-to-video experiments.}
    \vspace{-1mm}
    \label{tab:text2video_arch_base}
    \begin{tabular}{l c}
        \toprule
        \multicolumn{2}{c}{Text-to-video base model (1.08B parameters)}\\
        \midrule
        Channel multiplier & $[1, 2, 4, 4]$ \\
        Dropout & $0$ \\
        Number of channels & $256$ \\
        Number of residual blocks & $3$ \\
        Spatial self attention resolutions & $[32, 16, 8]$ \\
        Spatial cross attention resolutions & $[32, 16, 8]$ \\
        Temporal attention resolution & $[32, 16, 8]$ \\
        Number of channels in attention heads & 64\\
        Use scale shift norm & True \\
        \bottomrule
    \end{tabular}
\end{table}

\begin{table}[!ht]
    \centering
    \caption{Architecture for the temporal interpolation model in text-to-video experiments.}
    \vspace{-1mm}
    \label{tab:text2video_arch_ti}
    \begin{tabular}{l c}
        \toprule
        \multicolumn{2}{c}{Temporal interpolation model (1.08B parameters)}\\
        \midrule
        Channel multiplier & $[1, 2, 4, 4]$ \\
        Dropout & $0$ \\
        Number of channels & $256$ \\
        Number of residual blocks & $3$ \\
        Spatial self attention resolutions & $[32, 16, 8]$ \\
        Spatial cross attention resolutions & $[32, 16, 8]$ \\
        Temporal attention resolution & $[32, 16, 8]$ \\
        Number of channels in attention heads & 64\\
        Use scale shift norm & True \\
        \bottomrule
    \end{tabular}
\end{table}

\begin{table}[!ht]
    \centering
    \caption{Architecture for the spatial super-resolution model in text-to-video experiments.}
    \vspace{-1mm}
    \label{tab:text2video_arch_sr256}
    \begin{tabular}{l c}
        \toprule
        \multicolumn{2}{c}{Spatial super-resolution 256 (300M parameters)}\\
        \midrule
        Channel multiplier & $[1, 2, 4, 8]$ \\
        Block multiplier & $[1, 2, 4, 4]$ \\
        Dropout & $0$ \\
        Number of channels & $128$ \\
        Number of residual blocks & $2$ \\
        Spatial self attention resolutions & $[32]$ \\
        Spatial cross attention resolutions & $[32]$ \\
        Number of channels in attention heads & 64\\
        Use scale shift norm & True \\
        \bottomrule
    \end{tabular}
\end{table}

\begin{table}[!ht]
    \centering
    \caption{Architecture for the spatial super-resolution model in text-to-video experiments.}
    \vspace{-1mm}
    \label{tab:text2video_arch_sr1024}
    \begin{tabular}{l c}
        \toprule
        \multicolumn{2}{c}{Spatial super-resolution 1024 (170M parameters)}\\
        \midrule
        Patch size & $256 \times 256$ \\
        Channel multiplier & $[1, 2, 4, 4]$ \\
        Block multiplier & $[1, 2, 4, 4]$ \\
        Number of channels & $128$ \\
        Number of residual blocks & $2$ \\
        Spatial cross attention resolutions & $[32]$ \\
        Use scale shift norm & True \\
        \bottomrule
    \end{tabular}
\end{table}

\clearpage
{\small
\bibliographystyle{ieee_fullname}
\bibliography{egbib}

\begin{thebibliography}{10}\itemsep=-1pt

\bibitem{an2023latent}
Jie An, Songyang Zhang, Harry Yang, Sonal Gupta, Jia-Bin Huang, Jiebo Luo, and
  Xi Yin.
\newblock Latent-shift: Latent diffusion with temporal shift for efficient
  text-to-video generation.
\newblock {\em arXiv preprint arXiv:2304.08477}, 2023.

\bibitem{balaji2022ediffi}
Yogesh Balaji, Seungjun Nah, Xun Huang, Arash Vahdat, Jiaming Song, Qinsheng
  Zhang, Karsten Kreis, Miika Aittala, Timo Aila, Samuli Laine, Bryan
  Catanzaro, et~al.
\newblock {eDiff-I}: Text-to-image diffusion models with an ensemble of expert
  denoisers.
\newblock {\em arXiv preprint arXiv:2211.01324}, 2022.

\bibitem{bar2022text2live}
Omer Bar-Tal, Dolev Ofri-Amar, Rafail Fridman, Yoni Kasten, and Tali Dekel.
\newblock Text2live: Text-driven layered image and video editing.
\newblock In {\em ECCV}, pages 707--723. Springer, 2022.

\bibitem{blattmann2023videoldm}
Andreas Blattmann, Robin Rombach, Huan Ling, Tim Dockhorn, Seung~Wook Kim,
  Sanja Fidler, and Karsten Kreis.
\newblock Align your latents: High-resolution video synthesis with latent
  diffusion models.
\newblock In {\em CVPR}, 2023.

\bibitem{brooks2022instructpix2pix}
Tim Brooks, Aleksander Holynski, and Alexei~A Efros.
\newblock Instructpix2pix: Learning to follow image editing instructions.
\newblock {\em CVPR}, 2023.

\bibitem{carreira2017quo}
Joao Carreira and Andrew Zisserman.
\newblock Quo vadis, action recognition? a new model and the kinetics dataset.
\newblock In {\em CVPR}, pages 6299--6308, 2017.

\bibitem{ceylan2023pix2video}
Duygu Ceylan, Chun-Hao Huang, and Niloy~J. Mitra.
\newblock Pix2video: Video editing using image diffusion.
\newblock {\em arXiv:2303.12688}, 2023.

\bibitem{esser2023structure}
Patrick Esser, Johnathan Chiu, Parmida Atighehchian, Jonathan Granskog, and
  Anastasis Germanidis.
\newblock Structure and content-guided video synthesis with diffusion models.
\newblock {\em arXiv preprint arXiv:2302.03011}, 2023.

\bibitem{ge2022long}
Songwei Ge, Thomas Hayes, Harry Yang, Xi Yin, Guan Pang, David Jacobs, Jia-Bin
  Huang, and Devi Parikh.
\newblock Long video generation with time-agnostic vqgan and time-sensitive
  transformer.
\newblock {\em arXiv preprint arXiv:2204.03638}, 2022.

\bibitem{ge2023expressive}
Songwei Ge, Taesung Park, Jun-Yan Zhu, and Jia-Bin Huang.
\newblock Expressive text-to-image generation with rich text.
\newblock {\em arXiv preprint arXiv:2304.06720}, 2023.

\bibitem{gu2022vector}
Shuyang Gu, Dong Chen, Jianmin Bao, Fang Wen, Bo Zhang, Dongdong Chen, Lu Yuan,
  and Baining Guo.
\newblock Vector quantized diffusion model for text-to-image synthesis.
\newblock In {\em CVPR}, pages 10696--10706, 2022.

\bibitem{harvey2022flexible}
William Harvey, Saeid Naderiparizi, Vaden Masrani, Christian Weilbach, and
  Frank Wood.
\newblock Flexible diffusion modeling of long videos.
\newblock {\em arXiv preprint arXiv:2205.11495}, 2022.

\bibitem{hertz2022prompt}
Amir Hertz, Ron Mokady, Jay Tenenbaum, Kfir Aberman, Yael Pritch, and Daniel
  Cohen-Or.
\newblock Prompt-to-prompt image editing with cross attention control.
\newblock {\em arXiv preprint arXiv:2208.01626}, 2022.

\bibitem{ho2022imagen}
Jonathan Ho, William Chan, Chitwan Saharia, Jay Whang, Ruiqi Gao, Alexey
  Gritsenko, Diederik~P Kingma, Ben Poole, Mohammad Norouzi, David~J Fleet,
  et~al.
\newblock Imagen video: High definition video generation with diffusion models.
\newblock {\em arXiv preprint arXiv:2210.02303}, 2022.

\bibitem{ho2020denoising}
Jonathan Ho, Ajay Jain, and Pieter Abbeel.
\newblock Denoising diffusion probabilistic models.
\newblock {\em NeurIPS}, 33:6840--6851, 2020.

\bibitem{ho2022cascaded}
Jonathan Ho, Chitwan Saharia, William Chan, David~J Fleet, Mohammad Norouzi,
  and Tim Salimans.
\newblock Cascaded diffusion models for high fidelity image generation.
\newblock {\em JMLR}, 23:47--1, 2022.

\bibitem{ho2022video}
Jonathan Ho, Tim Salimans, Alexey Gritsenko, William Chan, Mohammad Norouzi,
  and David~J Fleet.
\newblock Video diffusion models.
\newblock {\em arXiv preprint arXiv:2204.03458}, 2022.

\bibitem{hong2022cogvideo}
Wenyi Hong, Ming Ding, Wendi Zheng, Xinghan Liu, and Jie Tang.
\newblock Cogvideo: Large-scale pretraining for text-to-video generation via
  transformers.
\newblock {\em arXiv preprint arXiv:2205.15868}, 2022.

\bibitem{hoppe2022diffusion}
Tobias H{\"o}ppe, Arash Mehrjou, Stefan Bauer, Didrik Nielsen, and Andrea
  Dittadi.
\newblock Diffusion models for video prediction and infilling.
\newblock {\em arXiv preprint arXiv:2206.07696}, 2022.

\bibitem{kahembwe2020lower}
Emmanuel Kahembwe and Subramanian Ramamoorthy.
\newblock Lower dimensional kernels for video discriminators.
\newblock {\em Neural Networks}, 2020.

\bibitem{karras2022elucidating}
Tero Karras, Miika Aittala, Timo Aila, and Samuli Laine.
\newblock Elucidating the design space of diffusion-based generative models.
\newblock {\em arXiv preprint arXiv:2206.00364}, 2022.

\bibitem{khachatryan2023text2video}
Levon Khachatryan, Andranik Movsisyan, Vahram Tadevosyan, Roberto Henschel,
  Zhangyang Wang, Shant Navasardyan, and Humphrey Shi.
\newblock Text2video-zero: Text-to-image diffusion models are zero-shot video
  generators.
\newblock {\em arXiv preprint arXiv:2303.13439}, 2023.

\bibitem{kumari2022multi}
Nupur Kumari, Bingliang Zhang, Richard Zhang, Eli Shechtman, and Jun-Yan Zhu.
\newblock Multi-concept customization of text-to-image diffusion.
\newblock {\em arXiv preprint arXiv:2212.04488}, 2022.

\bibitem{kynkaanniemi2023the}
Tuomas Kynk{\"a}{\"a}nniemi, Tero Karras, Miika Aittala, Timo Aila, and Jaakko
  Lehtinen.
\newblock The role of imagenet classes in fr\'echet inception distance.
\newblock In {\em ICLR}, 2023.

\bibitem{le2021ccvs}
Guillaume Le~Moing, Jean Ponce, and Cordelia Schmid.
\newblock Ccvs: Context-aware controllable video synthesis.
\newblock {\em NeurIPS}, 2021.

\bibitem{liu2022pseudo}
Luping Liu, Yi Ren, Zhijie Lin, and Zhou Zhao.
\newblock Pseudo numerical methods for diffusion models on manifolds.
\newblock {\em arXiv preprint arXiv:2202.09778}, 2022.

\bibitem{lu2022dpm}
Cheng Lu, Yuhao Zhou, Fan Bao, Jianfei Chen, Chongxuan Li, and Jun Zhu.
\newblock Dpm-solver++: Fast solver for guided sampling of diffusion
  probabilistic models.
\newblock {\em arXiv preprint arXiv:2211.01095}, 2022.

\bibitem{luo2023videofusion}
Zhengxiong Luo, Dayou Chen, Yingya Zhang, Yan Huang, Liang Wang, Yujun Shen,
  Deli Zhao, Jingren Zhou, and Tieniu Tan.
\newblock Videofusion: Decomposed diffusion models for high-quality video
  generation.
\newblock In {\em CVPR}, 2023.

\bibitem{ma2023follow}
Yue Ma, Yingqing He, Xiaodong Cun, Xintao Wang, Ying Shan, Xiu Li, and Qifeng
  Chen.
\newblock Follow your pose: Pose-guided text-to-video generation using
  pose-free videos.
\newblock {\em arXiv preprint arXiv:2304.01186}, 2023.

\bibitem{meng2022sdedit}
Chenlin Meng, Yutong He, Yang Song, Jiaming Song, Jiajun Wu, Jun-Yan Zhu, and
  Stefano Ermon.
\newblock {SDE}dit: Guided image synthesis and editing with stochastic
  differential equations.
\newblock In {\em ICLR}, 2022.

\bibitem{molad2023dreamix}
Eyal Molad, Eliahu Horwitz, Dani Valevski, Alex~Rav Acha, Yossi Matias, Yael
  Pritch, Yaniv Leviathan, and Yedid Hoshen.
\newblock Dreamix: Video diffusion models are general video editors.
\newblock {\em arXiv preprint arXiv:2302.01329}, 2023.

\bibitem{nichol2021glide}
Alex Nichol, Prafulla Dhariwal, Aditya Ramesh, Pranav Shyam, Pamela Mishkin,
  Bob McGrew, Ilya Sutskever, and Mark Chen.
\newblock Glide: Towards photorealistic image generation and editing with
  text-guided diffusion models.
\newblock {\em arXiv preprint arXiv:2112.10741}, 2021.

\bibitem{nikankin2022sinfusion}
Yaniv Nikankin, Niv Haim, and Michal Irani.
\newblock Sinfusion: Training diffusion models on a single image or video.
\newblock {\em arXiv preprint arXiv:2211.11743}, 2022.

\bibitem{parmar2023zero}
Gaurav Parmar, Krishna~Kumar Singh, Richard Zhang, Yijun Li, Jingwan Lu, and
  Jun-Yan Zhu.
\newblock Zero-shot image-to-image translation.
\newblock {\em arXiv preprint arXiv:2302.03027}, 2023.

\bibitem{qi2023fatezero}
Chenyang Qi, Xiaodong Cun, Yong Zhang, Chenyang Lei, Xintao Wang, Ying Shan,
  and Qifeng Chen.
\newblock Fatezero: Fusing attentions for zero-shot text-based video editing.
\newblock {\em arXiv preprint arXiv:2303.09535}, 2023.

\bibitem{radford2021learning}
Alec Radford, Jong~Wook Kim, Chris Hallacy, Aditya Ramesh, Gabriel Goh,
  Sandhini Agarwal, Girish Sastry, Amanda Askell, Pamela Mishkin, Jack Clark,
  Gretchen Krueger, and Ilya Sutskever.
\newblock Learning transferable visual models from natural language
  supervision.
\newblock In {\em ICML}, 2021.

\bibitem{raffel2020exploring}
Colin Raffel, Noam Shazeer, Adam Roberts, Katherine Lee, Sharan Narang, Michael
  Matena, Yanqi Zhou, Wei Li, and Peter~J. Liu.
\newblock Exploring the limits of transfer learning with a unified text-to-text
  transformer.
\newblock {\em JMLR}, 21(140):1--67, 2020.

\bibitem{ramesh2022hierarchical}
Aditya Ramesh, Prafulla Dhariwal, Alex Nichol, Casey Chu, and Mark Chen.
\newblock Hierarchical text-conditional image generation with clip latents.
\newblock {\em arXiv preprint arXiv:2204.06125}, 2022.

\bibitem{ranzato2014video}
MarcAurelio Ranzato, Arthur Szlam, Joan Bruna, Michael Mathieu, Ronan
  Collobert, and Sumit Chopra.
\newblock Video (language) modeling: a baseline for generative models of
  natural videos.
\newblock {\em arXiv preprint arXiv:1412.6604}, 2014.

\bibitem{rombach2022high}
Robin Rombach, Andreas Blattmann, Dominik Lorenz, Patrick Esser, and Bj{\"o}rn
  Ommer.
\newblock High-resolution image synthesis with latent diffusion models.
\newblock In {\em CVPR}, pages 10684--10695, 2022.

\bibitem{ruiz2022dreambooth}
Nataniel Ruiz, Yuanzhen Li, Varun Jampani, Yael Pritch, Michael Rubinstein, and
  Kfir Aberman.
\newblock Dreambooth: Fine tuning text-to-image diffusion models for
  subject-driven generation.
\newblock {\em arXiv preprint arXiv:2208.12242}, 2022.

\bibitem{saharia2022photorealistic}
Chitwan Saharia, William Chan, Saurabh Saxena, Lala Li, Jay Whang, Emily
  Denton, Seyed Kamyar~Seyed Ghasemipour, Burcu~Karagol Ayan, S~Sara Mahdavi,
  Rapha~Gontijo Lopes, et~al.
\newblock Photorealistic text-to-image diffusion models with deep language
  understanding.
\newblock {\em arXiv preprint arXiv:2205.11487}, 2022.

\bibitem{saito2017temporal}
Masaki Saito, Eiichi Matsumoto, and Shunta Saito.
\newblock Temporal generative adversarial nets with singular value clipping.
\newblock In {\em ICCV}, 2017.

\bibitem{saito2020train}
Masaki Saito, Shunta Saito, Masanori Koyama, and Sosuke Kobayashi.
\newblock Train sparsely, generate densely: Memory-efficient unsupervised
  training of high-resolution temporal gan.
\newblock {\em IJCV}, 2020.

\bibitem{Shen_2023_CVPR}
Xiaoqian Shen, Xiang Li, and Mohamed Elhoseiny.
\newblock Mostgan-v: Video generation with temporal motion styles.
\newblock In {\em CVPR}, 2023.

\bibitem{singer2022make}
Uriel Singer, Adam Polyak, Thomas Hayes, Xi Yin, Jie An, Songyang Zhang, Qiyuan
  Hu, Harry Yang, Oron Ashual, Oran Gafni, et~al.
\newblock Make-a-video: Text-to-video generation without text-video data.
\newblock {\em arXiv preprint arXiv:2209.14792}, 2022.

\bibitem{skorokhodov2021stylegan}
Ivan Skorokhodov, Sergey Tulyakov, and Mohamed Elhoseiny.
\newblock Stylegan-v: A continuous video generator with the price, image
  quality and perks of stylegan2.
\newblock {\em arXiv preprint arXiv:2112.14683}, 2021.

\bibitem{song2021denoising}
Jiaming Song, Chenlin Meng, and Stefano Ermon.
\newblock Denoising diffusion implicit models.
\newblock In {\em ICLR}, 2021.

\bibitem{song2021scorebased}
Yang Song, Jascha Sohl-Dickstein, Diederik~P Kingma, Abhishek Kumar, Stefano
  Ermon, and Ben Poole.
\newblock Score-based generative modeling through stochastic differential
  equations.
\newblock In {\em ICLR}, 2021.

\bibitem{soomro2012ucf101}
Khurram Soomro, Amir~Roshan Zamir, and Mubarak Shah.
\newblock Ucf101: A dataset of 101 human actions classes from videos in the
  wild.
\newblock {\em arXiv preprint arXiv:1212.0402}, 2012.

\bibitem{srivastava2015unsupervised}
Nitish Srivastava, Elman Mansimov, and Ruslan Salakhudinov.
\newblock Unsupervised learning of video representations using lstms.
\newblock In {\em ICML}, 2015.

\bibitem{tian2021a}
Yu Tian, Jian Ren, Menglei Chai, Kyle Olszewski, Xi Peng, Dimitris~N. Metaxas,
  and Sergey Tulyakov.
\newblock A good image generator is what you need for high-resolution video
  synthesis.
\newblock In {\em ICLR}, 2021.

\bibitem{tran2015learning}
Du Tran, Lubomir Bourdev, Rob Fergus, Lorenzo Torresani, and Manohar Paluri.
\newblock Learning spatiotemporal features with 3d convolutional networks.
\newblock In {\em ICCV}, 2015.

\bibitem{Tulyakov_2018_CVPR}
Sergey Tulyakov, Ming-Yu Liu, Xiaodong Yang, and Jan Kautz.
\newblock Mocogan: Decomposing motion and content for video generation.
\newblock In {\em CVPR}, June 2018.

\bibitem{unterthiner2018towards}
Thomas Unterthiner, Sjoerd van Steenkiste, Karol Kurach, Raphael Marinier,
  Marcin Michalski, and Sylvain Gelly.
\newblock Towards accurate generative models of video: A new metric \&
  challenges.
\newblock {\em ICLR}, 2019.

\bibitem{villegas2022phenaki}
Ruben Villegas, Mohammad Babaeizadeh, Pieter-Jan Kindermans, Hernan Moraldo,
  Han Zhang, Mohammad~Taghi Saffar, Santiago Castro, Julius Kunze, and Dumitru
  Erhan.
\newblock Phenaki: Variable length video generation from open domain textual
  description.
\newblock {\em arXiv preprint arXiv:2210.02399}, 2022.

\bibitem{voleti2022masked}
Vikram Voleti, Alexia Jolicoeur-Martineau, and Christopher Pal.
\newblock Masked conditional video diffusion for prediction, generation, and
  interpolation.
\newblock {\em arXiv preprint arXiv:2205.09853}, 2022.

\bibitem{vondrick2016generating}
Carl Vondrick, Hamed Pirsiavash, and Antonio Torralba.
\newblock Generating videos with scene dynamics.
\newblock {\em NIPS}, 2016.

\bibitem{wang2023videofactory}
Wenjing Wang, Huan Yang, Zixi Tuo, Huiguo He, Junchen Zhu, Jianlong Fu, and
  Jiaying Liu.
\newblock Videofactory: Swap attention in spatiotemporal diffusions for
  text-to-video generation.
\newblock {\em arXiv preprint arXiv:2305.10874}, 2023.

\bibitem{wu2022nuwa}
Chenfei Wu, Jian Liang, Lei Ji, Fan Yang, Yuejian Fang, Daxin Jiang, and Nan
  Duan.
\newblock N{\"u}wa: Visual synthesis pre-training for neural visual world
  creation.
\newblock In {\em ECCV}, pages 720--736. Springer, 2022.

\bibitem{wu2022tune}
Jay~Zhangjie Wu, Yixiao Ge, Xintao Wang, Weixian Lei, Yuchao Gu, Wynne Hsu,
  Ying Shan, Xiaohu Qie, and Mike~Zheng Shou.
\newblock Tune-a-video: One-shot tuning of image diffusion models for
  text-to-video generation.
\newblock {\em arXiv preprint arXiv:2212.11565}, 2022.

\bibitem{wu2018group}
Yuxin Wu and Kaiming He.
\newblock Group normalization.
\newblock In {\em ECCV}, pages 3--19, 2018.

\bibitem{xu2016msr}
Jun Xu, Tao Mei, Ting Yao, and Yong Rui.
\newblock Msr-vtt: A large video description dataset for bridging video and
  language.
\newblock In {\em CVPR}, 2016.

\bibitem{yan2021videogpt}
Wilson Yan, Yunzhi Zhang, Pieter Abbeel, and Aravind Srinivas.
\newblock Videogpt: Video generation using vq-vae and transformers.
\newblock {\em arXiv preprint arXiv:2104.10157}, 2021.

\bibitem{yang2022diffusion}
Ruihan Yang, Prakhar Srivastava, and Stephan Mandt.
\newblock Diffusion probabilistic modeling for video generation.
\newblock {\em arXiv preprint arXiv:2203.09481}, 2022.

\bibitem{yu2022magvit}
Lijun Yu, Yong Cheng, Kihyuk Sohn, Jos{\'e} Lezama, Han Zhang, Huiwen Chang,
  Alexander~G Hauptmann, Ming-Hsuan Yang, Yuan Hao, Irfan Essa, et~al.
\newblock Magvit: Masked generative video transformer.
\newblock {\em arXiv preprint arXiv:2212.05199}, 2022.

\bibitem{yu2021generating}
Sihyun Yu, Jihoon Tack, Sangwoo Mo, Hyunsu Kim, Junho Kim, Jung-Woo Ha, and
  Jinwoo Shin.
\newblock Generating videos with dynamics-aware implicit generative adversarial
  networks.
\newblock In {\em ICLR}, 2021.

\bibitem{zhang2023adding}
Lvmin Zhang and Maneesh Agrawala.
\newblock Adding conditional control to text-to-image diffusion models.
\newblock {\em arXiv preprint arXiv:2302.05543}, 2023.

\bibitem{zhang2023fast}
Qinsheng Zhang and Yongxin Chen.
\newblock Fast sampling of diffusion models with exponential integrator.
\newblock In {\em ICLR}, 2023.

\bibitem{zhou2022magicvideo}
Daquan Zhou, Weimin Wang, Hanshu Yan, Weiwei Lv, Yizhe Zhu, and Jiashi Feng.
\newblock Magicvideo: Efficient video generation with latent diffusion models.
\newblock {\em arXiv preprint arXiv:2211.11018}, 2022.

\end{thebibliography}
}

\end{document}